\documentclass[lettersize,journal]{IEEEtran}
\usepackage{amsmath,amsfonts}
\usepackage{algorithmic}
\usepackage{algorithm}
\usepackage{array}
\usepackage[caption=false,font=normalsize,labelfont=sf,textfont=sf]{subfig}
\usepackage{textcomp}
\usepackage{stfloats}
\usepackage{url}
\usepackage{verbatim}
\usepackage{graphicx}
\usepackage{amssymb}
\usepackage{cite}
\usepackage{pgfplots}
\usepackage{multirow}
\usepackage{booktabs}
\usepackage{marvosym}
\usepackage{mathtools}
\usepackage{tikz, flushend}
\pgfplotsset{compat=1.18}
\DeclareMathSymbol{\mathbbE}{\mathord}{AMSb}{"45}

\newcommand{\ex}{\mathbbE}
\usepackage{pgfplotstable}
\newcommand*\circled[1]{\tikz[baseline=(char.base)]{
            \node[shape=circle,draw,inner sep=0.8pt] (char) {#1};}}
            
\begin{document}

\title{A Dual-Purpose Framework for Backdoor Defense and Backdoor Amplification in Diffusion Models}

\author{Vu Tuan Truong,~\IEEEmembership{Student Member,~IEEE,} Long Bao Le,~\IEEEmembership{Fellow,~IEEE}
\thanks{T. V. Truong and L. B. Le are with INRS, University of Qu\'{e}bec, Montr\'{e}al, QC H5A 1K6, Canada (email: tuan.vu.truong@inrs.ca, long.le@inrs.ca).}
}



\maketitle

\begin{abstract}
Diffusion models have emerged as state-of-the-art generative frameworks, excelling in producing high-quality multi-modal samples. However, recent studies have revealed their vulnerability to backdoor attacks, where backdoored models generate specific, undesirable outputs called backdoor target (e.g., harmful images) when a pre-defined trigger is embedded to their inputs. In this paper, we propose PureDiffusion, a dual-purpose framework that simultaneously serves two contrasting roles: backdoor defense and backdoor attack amplification. For defense, we introduce two novel loss functions to invert backdoor triggers embedded in diffusion models. The first leverages trigger-induced distribution shifts across multiple timesteps of the diffusion process, while the second exploits the denoising consistency effect when a backdoor is activated. Once an accurate trigger inversion is achieved, we develop a backdoor detection method that analyzes both the inverted trigger and the generated backdoor targets to identify backdoor attacks.
In terms of attack amplification with the role of an attacker, we describe how our trigger inversion algorithm can be used to reinforce the original trigger embedded in the backdoored diffusion model. This significantly boosts attack performance while reducing the required backdoor training time.
Experimental results demonstrate that PureDiffusion achieves near-perfect detection accuracy, outperforming existing defenses by a large margin, particularly against complex trigger patterns. Additionally, in an attack scenario, our attack amplification approach elevates the attack success rate (ASR) of existing backdoor attacks to nearly 100\% while reducing training time by up to 20×.
\end{abstract}


\section{Introduction}\label{section:introduction}

In recent years, diffusion models have rapidly risen to prominence, establishing a new state-of-the-art across various categories of deep generative models. Leveraging a multi-step denoising approach~\cite{ho2020denoising}, diffusion models have demonstrated exceptional performance in diverse generative tasks, including computer vision~\cite{watson2021learning, nichol2021glide, sinha2021d2c, bao2022analytic}, natural language processing (NLP)~\cite{austin2021structured,  hoogeboom2021argmax, li2022diffusion, savinov2021step}, 3D synthesis~\cite{xu2023dream3d, truong2024text, poole2022dreamfusion}, audio processing~\cite{chen2020wavegrad, popov2021grad}, and time series~\cite{tashiro2021csdi, yan2021scoregrad, rasul2020multivariate}. These models consistently outperform earlier methods such as generative adversarial networks (GANs)~\cite{goodfellow2014generative} and variational autoencoders (VAEs)~\cite{kingma2014vae, rezende2015variational}. The recent development of diffusion models is rapid, introducing various categories of models such as denoising diffusion probabilistic models (DDPMs)~\cite{ho2020denoising}, denoising diffusion implicit models (DDIMs)~\cite{song2020denoising}, noise conditioned score networks (NCSNs)~\cite{song2019generative}, stochastic differential equation (SDE)\cite{song2020score}, and ordinary differential equation (ODE).

However, recent studies showed the vulnerability of diffusion models to backdoor attacks~\cite{truong2024attacks}. In traditional classification tasks, backdoor attacks are typically performed by poisoning a portion of the training data with a specific trigger and assigning it a target label~\cite{li2022backdoor}. A classifier trained on this poisoned dataset will output the target label when the trigger is present in its input, while behaving normally otherwise to maintain stealthiness and utility. 
On the other hand, diffusion models operate differently: they take Gaussian noise as input and perform generative tasks rather than classification~\cite{ho2020denoising}. Backdoor attacks on diffusion models involve fine-tuning a clean model so that it generates a specific undesirable target (e.g., harmful or copyright-infringing content) when the input noise is stamped with a predefined trigger~\cite{truong2024attacks}. However, when clean noise is provided, the model continues to produce high-quality, benign outputs, ensuring the attack remains covert.

The widespread availability of diffusion models on open platforms (e.g., Hugging Face) poses significant risks, as backdoor attacks could result in severe consequences for both organizations and individual users. For instance, a startup might unknowingly download a compromised diffusion model and integrate it into their generative AI services. If the backdoor is triggered, the model could generate harmful content for paying customers or produce outputs that violate copyright laws, exposing the company to both ethical and legal challenges. These risks are further amplified when backdoor-affected diffusion models are used in downstream applications such as adversarial purification or robustness certification, potentially compromising entire systems even when the diffusion model is only a single component.

Despite these threats, defense mechanisms against backdoor attacks in diffusion models remain largely underexplored. Most existing backdoor detection frameworks are designed for traditional classification models~\cite{wang2019neural, chen2018detecting} and cannot be directly applied to generative diffusion models. Detecting backdoors in diffusion models poses unique challenges: gradient-based methods struggle due to the models’ reliance on thousands of denoising steps, while search-based approaches are nearly infeasible because the outputs reside in a vast generative space (e.g., high-resolution image space) rather than being confined to a finite set of classes. 

In order to protect diffusion models from backdoor attacks, several prior studies have been conducted~\cite{an2024elijah, guan2024ufid, sui2024disdet, hao2024diff, mo2024terd}. However, they often require infeasible assumptions that hinder them from real-world scenarios. For instance, DisDet~\cite{sui2024disdet} and UFID~\cite{guan2024ufid} assume the availability of a predefined set of candidate triggers, which is used to assess whether a diffusion model is backdoored. In practice, defenders typically lack such information and must rely on a complex process called trigger inversion to identify candidate triggers, which remains one of the most challenging tasks in this domain. To date, only Elijah~\cite{an2024elijah} and TERD~\cite{mo2024terd} incorporate trigger inversion into their frameworks, making them better suited for real-world applications. However, as demonstrated in our evaluations (Section~\ref{section:experiment}), these methods perform effectively only in standard cases where the backdoored models have a high poisoning rate and easily identifiable trigger patterns. In more difficult scenarios, such as when backdoored models have a low poisoning rate or involve complex, hard-to-invert triggers, both frameworks suffer from low detection accuracy and often fail to successfully invert the triggers.

To bridge the gap, we propose PureDiffusion, a novel backdoor defense framework for diffusion models consisting of two key steps: trigger inversion and backdoor detection. Our trigger inversion mechanism leverages two critical backdoor traces: \circled{1} \textbf{Trigger Shift:} The denoising process in backdoored diffusion models introduces a trigger-related distribution shift (``trigger shift") that gradually removes the trigger pattern from the input noise over successive denoising steps. By the final timestep, the generated output no longer contains the backdoor trigger. To exploit this phenomenon, we design the multi-timestep distribution-shift loss ($L_{MDS}$), which learns a trigger pattern that induces this trigger shift across all timesteps. \circled{2} \textbf{Denoising Consistency Effect:} Normally, diffusion models produce diverse outputs due to random sampling of input noise. However, if the input noise contains a stamped trigger, backdoored models consistently generate the same backdoor target. This behavior, referred to as ``target shift," adds the backdoor target incrementally during the denoising process until the backdoor target is fully reconstructed at the final timestep. To capture this effect, we introduce the denoising-consistency loss ($L_{DC}$), which learns a trigger pattern that creates similar distribution shifts between two denoising processes starting from different input noises. 
Following trigger inversion, our backdoor detection method determines whether a diffusion model is backdoored using both input- and output-based analyses. For input-based detection, we compare the distribution of the inverted trigger with that of benign distributions to assess the likelihood of an attack. For output-based detection, we evaluate the consistency of generated samples in the presence of the inverted trigger. High consistency across different trials suggests a higher likelihood of backdooring.

It is worth noting that Elijah~\cite{an2024elijah} also utilizes trigger shifts in their trigger inversion method, but their approach is limited to the first denoising timestep. This limitation arises because, while it is known that each denoising timestep removes a portion of the trigger pattern, the exact amount removed at each step remains uncomputable. In this paper, we address this gap by proposing three approaches to compute the scale of trigger shifts across multiple timesteps, tailored to different defense scenarios: white-box and gray-box settings. As a result, we extend the trigger-shift loss function to incorporate multiple denoising timesteps, significantly enhancing the robustness of trigger inversion.

Although our framework is designed for backdoor defense, our trigger inversion method can also serve as a trigger reinforcement mechanism, amplifying the effectiveness of backdoor attacks. Specifically, after injecting a backdoor into a diffusion model using a predefined trigger and target, attackers can apply our trigger reinforcement method to refine the original trigger. This refined trigger not only achieves a significantly higher attack success rate (ASR) but is also more imperceptible due to its altered shape. Furthermore, our trigger reinforcement technique is computationally efficient, requiring far less training effort than traditional backdoor attack methods. By leveraging this approach, attackers can reduce the number of backdoor training epochs by 10-20 times while still achieving higher ASR compared to using the original backdoor triggers.

To summarize, our contributions are as follows:

\begin{itemize}
    \item We propose a novel trigger inversion algorithm that efficiently learns backdoor triggers across multiple denoising timesteps, leveraging two essential backdoor traces. Our method achieves high attack success rates (ASR) and even outperforms ground-truth triggers in certain cases.
    \item We develop a robust backdoor detection mechanism that combines input-triggered noise analysis and output sample evaluation to identify manipulated diffusion models.
    \item  From an attacker’s perspective, we introduce a trigger reinforcement technique that enhances standard backdoor attacks by increasing ASR while reducing training time dramatically.
    \item We conduct extensive experiments to demonstrate the effectiveness of our dual-purpose framework across various scenarios, including different backdoor methods and diverse trigger-target combinations.
\end{itemize}

The rest of our paper is structured as follows. Section~\ref{section:related-work} reviews related work regarding backdoor attacks and defenses in diffusion models. Section~\ref{section:preliminary} provides background knowledge of diffusion models and backdoors. Section~\ref{section:methodology} describes our framework PureDiffusion. Section~\ref{section:experiment} shows the experimental results, while section~\ref{section:conclusion} concludes the paper.

\section{Related Work}\label{section:related-work}

\subsection{Backdoor Attacks in Diffusion Models}
While early diffusion models only contain a UNet~\cite{ronneberger2015u} for image generation, recent state-of-the-art diffusion models like stable diffusion further integrate conditional models (e.g., text encoders) connected to the UNet via cross-attention layers, enabling multi-modal constraints. Among the components, backdooring the UNet remains the most fundamental attack strategy, as explored in three notable studies: BadDiffusion~\cite{Chou2023CVPR}, TrojDiff~\cite{chen2023trojdiff}, and VillanDiffusion~\cite{chou2024villandiffusion}. 
To backdoor diffusion models, both BadDiffusion~\cite{Chou2023CVPR} and TrojDiff~\cite{chen2023trojdiff} rely on poisoning a subset of the training dataset, altering the diffusion process, and modifying the training objective. This approach gradually embeds a trigger into the prior noise distribution across multiple diffusion steps. Once backdoor training is complete, images generated from trigger-stamped noise become the desired target images. The key distinction between these two frameworks lies in their trigger schedules (i.e., parameters that control the scale of the trigger added at each diffusion step), resulting in different performance outcomes and attack scenarios. 
VillanDiffusion~\cite{chou2024villandiffusion}, on the other hand, extends the capabilities of BadDiffusion by generalizing backdoor attacks to various samplers (e.g., ODE and SDE samplers) and diffusion model categories (e.g., DDPM, DDIM, score-based models, and conditional models). This is achieved using numerical methods to compute a unified diffusion transition, allowing the formulation of generalizable attack strategies rather than relying on specific settings.
Despite these advances, all the presented backdoor attacks operate under black-box scenarios. They require customization for each target model and assume prior knowledge of the noise schedule, which limits their applicability and generalizability.

Several studies have explored backdooring diffusion models by targeting conditional models instead of the UNet. For instance, the authors in~\cite{struppek2023rickrolling} use non-Latin characters (e.g., \Smiley{}) as backdoor triggers, while the target is a specific text prompt such as ``a cat wearing glasses". By misleading the text embeddings to a specific target prompt before feeding them to diffusion models via cross-attention layers, the final generation results of diffusion models are manipulated correspondingly. Some other works also target on the textual modality to backdoor diffusion models~\cite{zhai2023text, pan2023trojan, wang2023stronger}. In general, they all share the same principle of poisoning a subset of the training data with a textual trigger and an associated backdoor target, differing primarily in how the trigger and target are selected. 
Essentially, in these backdoor attacks, attackers only manipulate the training dataset without interfering in the diffusion processes and the training objectives.

\subsection{Backdoor Defenses for Diffusion Models}
Backdoor defense mechanisms for diffusion models typically involve two main stages: trigger inversion and backdoor detection. Trigger inversion is the first and foremost stage, in which suspicious diffusion models are examined based on essential diffusion properties to identify one or more candidate backdoor triggers. Typically, the trigger is often a learnable value which is learned via optimization techniques\cite{truong2024attacks}. Backdoor detection forms the second stage, where the candidate triggers and the outputs generated using these triggers are analyzed to determine whether the suspicious models have been backdoored. 

Elijah~\cite{an2024elijah} is the first backdoor defense framework that offers both trigger inversion and backdoor detection. For trigger inversion, it identifies a candidate trigger by analyzing the distribution shift at the first denoising step. Backdoor detection is performed by computing the total variance loss of generated samples and the absolute distance between them in the presence of the inverted trigger to determine if the model is backdoored. The authors argue that a model is more likely to be backdoored if both the total variance score and absolute distance are low.
However, our experiments show that Elijah's trigger inversion method becomes inefficient when handling difficult trigger patterns. This limitation arises because the framework focuses solely on the first denoising step, missing critical information about the distribution shifts that occur across the other steps.
Another work named TERD\cite{mo2024terd} proposed a different approach for trigger inversion, which contains two steps: trigger estimation and trigger refinement. Trigger estimation is based on a loss function that compares the difference between denoising results from two different input noises in the presence of the learnable trigger. Trigger refinement technique uses the estimated triggers to generate potential backdoor targets. Then, the authors assume that the generated potential targets are such the true targets, thus using them to improve the quality of the estimated triggers. Regarding backdoor detection, instead of analyzing samples generated by the inverted trigger as in Elijah, TERD analyzes the inverted trigger itself to determine if a model is backdoored. However, TERD's trigger inversion results in low performance for difficult trigger-target pairs, while its backdoor detection could be bypassed if the trigger is intentionally constrained to resemble a normal noise\cite{li2023learnable}. On the other hand, our framework PureDiffusion offers high performance for both trigger inversion and backdoor detection even with challenging triggers which could not be detected by both Elijah and TERD. 

DisDet~\cite{sui2024disdet} and UFID~\cite{guan2024ufid} are only designed for backdoor detection, neglecting the trigger inversion stage. This means that both framework assume that there are already candidate triggers to be validated, which is an impractical assumption. Other studies such as T2Ishield~\cite{wang2025t2ishield} and CopyrightShield~\cite{guo2024copyrightshield} focus on defending conditional diffusion models by analyzing the text encoders and attention layers. Therefore, these framework cannot resist backdoor attacks targeting on the UNet model, which is the main goal of our work.

\subsection{Discussion}
Our work primarily serves as a defense against backdoor attacks targeting denoising UNet models. To the best of our knowledge, only Elijah \cite{an2024elijah} and TERD \cite{mo2024terd} have addressed this issue with practical solutions. However, both frameworks perform well only on default triggers (e.g., a white square box) and struggle with more complex ones.

Preliminary results of this work were published in \cite{Vu2025ICC}; however, this journal version makes several significant extensions compared to the conference version as can be summarized as follows. First, analysis of the trigger shifts  for three different backdoor attacks, namely BadDiffusion \cite{Chou2023CVPR}, TrojDiff \cite{chen2023trojdiff}, and VillanDiffusion \cite{chou2024villandiffusion} are performed in this version while only the analysis for BadDiffusion is conducted in \cite{Vu2025ICC}. Second, while we employ only a single loss function in  \cite{Vu2025ICC}, two different loss functions are employed for trigger inversion in this version for enhanced performance. Third, much more extensive experimental studies and results are presented in this version where we study various diffusion models (both DDPM-based and score-based diffusion models). 
Finally, we conduct the ablation study, complexity, and sensitivity analysis in this version, which are not pursued in \cite{Vu2025ICC}.


Our study focuses exclusively on backdoor attacks targeting the UNet, without considering attacks on other components, such as modality encoders (e.g., text encoders). 

\section{Preliminary}\label{section:preliminary}

\subsection{Background of Diffusion Models}
Diffusion models are deep generative models designed to generate high-quality samples via two key processes, which are the forward (diffusion) process, and the backward (denoising) process. In early models like DDPMs, these processes are represented as a Markov chain~\cite{ho2020denoising}. The forward process gradually adds Gaussian noise to clean images over a total of $T$ timesteps until the images are fully destroyed. This is achieved via the following forward transition: 
\begin{equation}
\label{equation:ddpm-forward}
    q(\mathbf{x}_t|\mathbf{x}_{t-1}) = \mathcal{N}(\mathbf{x}_t;\sqrt{\alpha_t}\mathbf{x}_{t-1}, (1-\alpha_t)\mathbf{I}),
\end{equation}
where $\alpha_t \in (0,1)$ controls the amount of noise added at each timestep, and $t \in (0;T]$ represents the current timestep. Using reparameterization, $\mathbf{x}_t$ can be directly sampled from the clean data $\mathbf{x}_0$ as:
\begin{equation}
\label{equation:direct-sampling}
\mathbf{x}_t = \sqrt{\Bar{\alpha}_t}\mathbf{x}_0 + \sqrt{1-\Bar{\alpha}_t}\boldsymbol{\epsilon}_0, 
\end{equation}
where $\Bar{\alpha}_t = \prod_{i=1}^{t} \alpha_{i}$ and $\boldsymbol{\epsilon}_0 \sim \mathcal{N}(0,\mathbf{I})$.

From this forward process, the ground-truth backward transition can be derived using Bayes' rule: $q(\mathbf{x}_{t-1}|\mathbf{x}_t,\mathbf{x}_0) = \frac{q(\mathbf{x}_t|\mathbf{x}_{t-1},\mathbf{x}_0)q(\mathbf{x}_{t-1}|\mathbf{x}_0)}{q(\mathbf{x}_{t}|\mathbf{x}_0)}.$ A deep neural network ${\boldsymbol{\theta}}$, often a UNet, is trained to approximate the backward transition:
\begin{equation}
\label{equation:backward-transition}
    p_{\boldsymbol{\theta}}(\mathbf{x}_{t-1}|\mathbf{x}_t) = \mathcal{N}(\mathbf{x}_{t-1}; \mu_{\boldsymbol{\theta}}(\mathbf{x}_t,t), \Sigma_{\boldsymbol{\theta}}(\mathbf{x}_t,t)),
\end{equation}
where  $\mu_{\boldsymbol{\theta}}(\mathbf{x}_t,t)$ and $\Sigma_{\boldsymbol{\theta}}(\mathbf{x}_t,t)$ are the predicted mean and variance of the image distribution. The network is trained by minimizing the KL divergence between the predicted backward transition $p_{\boldsymbol{\theta}}(\mathbf{x}_{t-1}|\mathbf{x}_t)$ and the ground-truth transition $q(\mathbf{x}_{t-1}|\mathbf{x}_t, \mathbf{x}_0)$, resulting in a simplified loss function:
\begin{equation}
\label{equation:loss-ddpm}
    \mathcal{L} = \ex_{\mathbf{x}_0,\boldsymbol{\epsilon}} \left[ \left\| \boldsymbol{\epsilon} - \boldsymbol{\epsilon}_{\boldsymbol{\theta}}(\mathbf{x}_t,t) \right\|^2_2 \right],
\end{equation}
where $\boldsymbol{\epsilon} \sim \mathcal{N}(0,\mathbf{I})$ and $\boldsymbol{\epsilon}_{\boldsymbol{\theta}}(\mathbf{x}_t,t)$ is the noise predicted by the UNet. 
Once trained, the UNet ${\boldsymbol{\theta}}$ can generate images from arbitrary Gaussian noise through the following denoising transition:
\begin{equation}
\label{equation:ddpm-sampling}
    \mathbf{x}_{t-1} = \frac{1}{\sqrt{\alpha_t}} \left( \mathbf{x}_t - \frac{1-\alpha_t}{\sqrt{1-\Bar{\alpha}_t}}\boldsymbol{\epsilon}_{\boldsymbol{\theta}}(\mathbf{x}_t,t) \right) + \sigma_t\boldsymbol{\epsilon},
\end{equation}
where $\sigma_t = \frac{(1-\alpha_t)(1-\Bar{\alpha}_{t-1})}{1-\Bar{\alpha}_{t}}$. 

However, a significant limitation of DDPMs is their slow sampling process, as the denoising procedure requires thousands of steps. To address this, Song et al.~\cite{song2020denoising} generalized DDPMs to non-Markovian diffusion chains, resulting in DDIMs. DDIMs dramatically reduce the number of denoising steps while maintaining high sample quality.
Moreover, whereas DDPMs and DDIMs are discrete in nature, Song et al.~\cite{song2020score} further extended diffusion models by formulating the process as a continuous-time SDE or ODE. This formulation allows for higher-order approximations, enabling high-quality sample generation with fewer steps.

\subsection{Diffusion Model Backdoors}
Backdoor attacks aim to make diffusion models produce a predefined backdoor target $\mathbf{x}_0^*$ (e.g., a harmful image) if the backdoor trigger ${\boldsymbol{\delta}}$ (e.g., a small white box) is attached to the input noise while sampling. We use ``*" to indicate backdoor. While the benign forward process only adds some Gaussian noise to clean images $\mathbf{x}_0$ in each step, the backdoored process adds not only the noise, but also a certain amount of trigger ${\boldsymbol{\delta}}$ to the data distribution. 
We present a unified formula for backdoor diffusion transition as follows:
\begin{equation}
\label{equation:backdoor-forward1}
    q(\mathbf{x}^*_t|\mathbf{x}^*_{t-1}) = \mathcal{N}(\mathbf{x}^*_t;a(t)\mathbf{x}^*_{t-1} + b(t){\boldsymbol{\delta}}, c(t)\mathbf{I}),
\end{equation}
where $a(t)$ is ``content schedulers", which determines the amount of image distribution kept from the previous timestep. The terms $b(t)$ and $c(t)$ are ``trigger schedulers" and ``noise schedulers", respectively, which control the amount of trigger and noise added to image distribution at each timestep. It can be seen that if we choose $a(t)=\sqrt{\alpha_t}$, $b(t)=0$, and $c(t)=1-\alpha_t$, the equation (\ref{equation:backdoor-forward1}) becomes the equation (\ref{equation:ddpm-forward}), which is the forward transition of a benign DDPM.

Each backdoor method results in different sets of $a(t)$, $b(t)$, and $c(t)$. However, these coefficients must be chosen subtly to enable the direct sampling of $\mathbf{x}^*_t$ from the backdoor target $\mathbf{x}^*_0$, which is similar to the property in the equation (\ref{equation:direct-sampling}). In the follows, we present three state-of-the-art backdoor attacks~\cite{Chou2023CVPR, chen2023trojdiff, chou2024villandiffusion} based on the above unified forward process.

\textbf{BadDiffusion.} The authors in~\cite{Chou2023CVPR} choose $a(t)=\sqrt{\alpha_t}$, $b(t)=1-\sqrt{\alpha_t}$, and $c(t)=1-\alpha_t$, resulting in the following forward transition and forward direct sampling:
\begin{equation}
\label{equation:backdoor-forward-baddiffusion}
    q(\mathbf{x}^*_t|\mathbf{x}^*_{t-1}) = \mathcal{N}(\mathbf{x}^*_t;\sqrt{\alpha_t}\mathbf{x}^*_{t-1} + (1-\sqrt{\alpha_t}){\boldsymbol{\delta}}, (1-\alpha_t)\mathbf{I}).
\end{equation}
\begin{equation}
\label{equation:baddifusion-direct-sampling}
    \mathbf{x}_t = \sqrt{\Bar{\alpha}_t}\mathbf{x}^*_0 + (1-\sqrt{\Bar{\alpha}_t}){\boldsymbol{\delta}} + \sqrt{1-\Bar{\alpha}_t}\boldsymbol{\epsilon}_0.
\end{equation}

\textbf{TrojDiff.} This attack selects $a(t)=\sqrt{\alpha_t}$, $b(t)=k_t(1-\gamma)$, and $c(t)=\gamma(1-\alpha_t)$, where $\gamma \in [0,1]$ is a blending coefficient that controls the scale of trigger added to the image distribution in each step, and $k_t$ is subtly chosen for every timestep $t$ to enable the direct sampling property. As a result, the forward process of TrojDiff is:
\begin{equation}
\label{equation:backdoor-forward-trojdiff}
    q(\mathbf{x}^*_t|\mathbf{x}^*_{t-1}) = \mathcal{N}(\mathbf{x}^*_t;\sqrt{\alpha_t}\mathbf{x}^*_{t-1} + k_t(1-\gamma){\boldsymbol{\delta}}, \gamma(1-\alpha_t)\mathbf{I}).
\end{equation}
\begin{equation}
\label{equation:trojdiff-direct-sampling}
    \mathbf{x}^*_t = \sqrt{\Bar{\alpha}_t}\mathbf{x}^*_0 + \sqrt{1-\Bar{\alpha}_t}(1-\gamma){\boldsymbol{\delta}} + \sqrt{1-\Bar{\alpha}_t}\gamma\boldsymbol{\epsilon}.
\end{equation}

\textbf{VillanDiffusion.} The authors in~\cite{chou2024villandiffusion} obtained $a(t)=v_t$, $b(t)=h_t$, and $c(t)=\sqrt{w_t}$ as unified formulas derived from the content scheduler $\Bar{\alpha}(t)$, noise scheduler $\Bar{\beta}(t)$, and a \textit{correction term} $\Bar{\rho}(t)$. Thus, the forward transition and forward sampling process of VillanDiffusion are:
\begin{equation}
\label{equation:backdoor-forward-villandiffusion}
    q(\mathbf{x}^*_t|\mathbf{x}^*_{t-1}) = \mathcal{N}(\mathbf{x}^*_t;v_t\mathbf{x}^*_{t-1} + h_t{\boldsymbol{\delta}}, w_t\mathbf{I}).
\end{equation}
\begin{equation}
\label{equation:villandiffusion-direct-sampling}
    \mathbf{x}^*_t =\Bar{\alpha}(t)\mathbf{x}^*_0 + \Bar{\rho}(t){\boldsymbol{\delta}} + \Bar{\beta}^2(t)\boldsymbol{\epsilon}.
\end{equation}

The details of VillanDiffusion's coefficients can be found in~\cite{chou2024villandiffusion}. Note that in the original manuscript, the authors denote $a(t)=k_t$. In our paper, we use $v_t$ to distinguish it from the $k_t$ coefficient of TrojDiff.

Based on the three presented backdoor forward processes, the corresponding backdoor backward processes can be derived to train backdoor models. In practice, attackers only apply the backdoor training process on a small portion of the dataset, while the rest of the training process is trained with benign diffusion process to ensure utility.

\begin{figure*}[h!]
	\centering
	\includegraphics[scale=0.15]{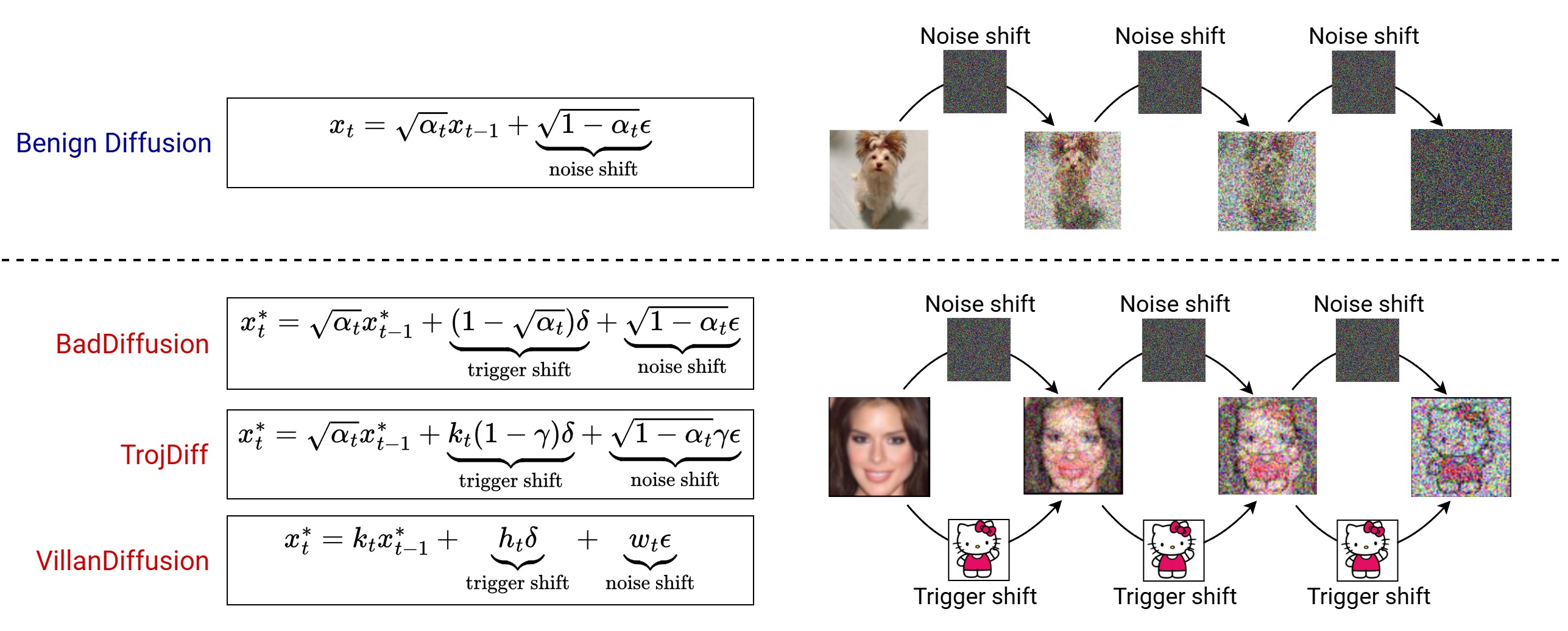}
	\caption{A visualization of distribution shifts caused by the forward diffusion process in both benign and backdoor scenarios.}
	\label{figure:distribution-shift}
\end{figure*}

\section{PureDiffusion Framework}\label{section:methodology}

\subsection{Threat Model}

We considered two entities in our attack and defenses settings, including attacker and defender.

\textbf{Attacker.} An attacker can conduct backdoor attack on a diffusion model using a predefined combination of backdoor trigger and target. 
Then, the attacker can upload the manipulated model to open platform like Hugging Face, Github, or cloud service. We assume that users can download the model to use locally or call API request to the cloud to receive the generation results from the cloud. 

\textbf{Defender.} The role of defender is to analyze diffusion models downloaded from open platforms, detecting if any model was backdoored. The defender only has access to the diffusion models and has no knowledge of the attacks (e.g., information about trigger and target).

Regarding defense settings, we consider two scenarios:

\textbf{White-Box Defense.} In this setting, the defender has access to the model's parameters (e.g., the UNet's weights) and other diffusion coefficients (e.g., the noise schedule $\alpha$), while it also control all the model's input, output, and denoising result in each step.

\textbf{Gray-Box Defense.} The defender has no access to the model's parameters and coefficients. they can only obtain the input noise, the output image, and intermediate denoising result in every step.

\begin{figure}[h!]
	\centering
	\includegraphics[scale=0.17]{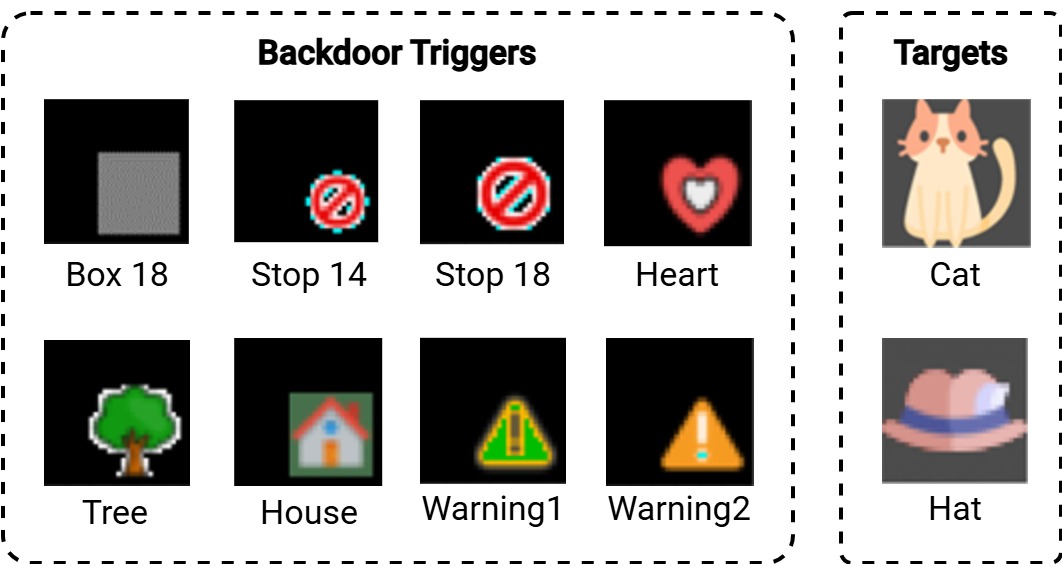}
	\caption{The backdoor triggers and targets chosen in our experiments.}
	\label{figure:backdoor-settings}
\end{figure}

Fig.~\ref{figure:backdoor-settings} illustrates different backdoor triggers and targets used in our paper.
Aside from utilizing diffusion models, we assume the option to fine-tune these models on a specific dataset provided by users is also available. 

\subsection{Backdoor Trigger Inversion}\label{subsection:trigger-inversion}

When backdoor targets are known, inverting the corresponding triggers using gradient-based methods is relatively straightforward. However, in real-world attacks, such information is rarely available, making trigger inversion a highly challenging task. In traditional classification tasks, backdoor targets are often confined to a limited number of classes, enabling defenders to exhaustively test each class as a potential target. In contrast, this strategy is infeasible for diffusion models due to their vast and diverse output space, inherent to generative tasks.

The unique properties arising from the diffusion processes can be leveraged to facilitate the inversion of backdoor triggers. Based on these properties, we propose two learning objectives for trigger inversion, namely $L_{MDS}$ and $L_{DC}$.

\begin{figure*}[h!]
	\centering
	\includegraphics[scale=0.13]{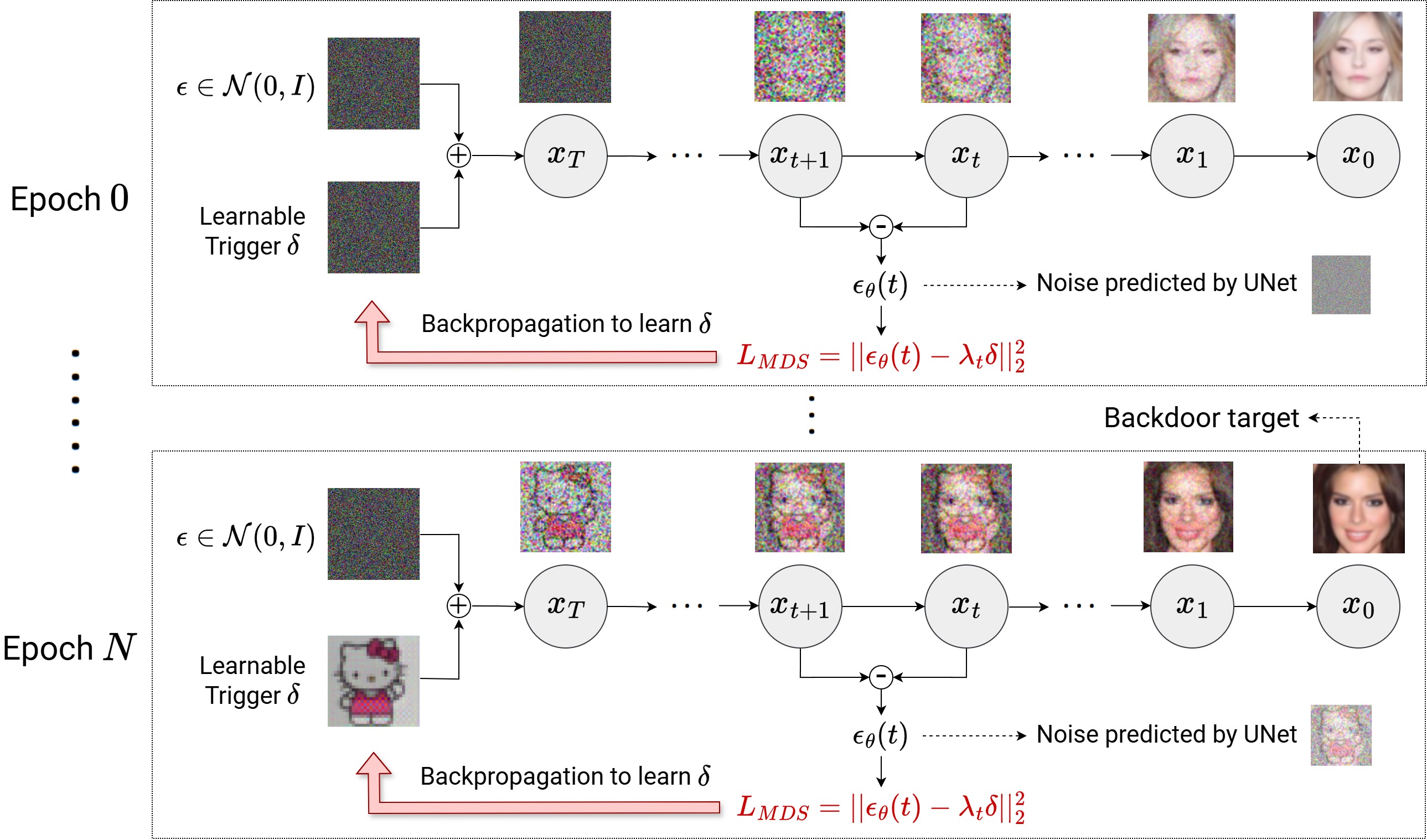}
	\caption{The training process using our multi-timestep distribution-shift loss $L_{MDS}$.}
	\label{figure:MDS-loss}
\end{figure*}

\subsubsection{Backdoor Traces from Diffusion Process}
By reparameterizing the backdoor forward transition in the equation (\ref{equation:backdoor-forward1}), the backdoor diffusion process can be interpreted as a process of gradually adding some noise and a certain amount of the backdoor trigger into the backdoor target:
\begin{equation}
\label{equation:backdoor-reparameterization}
    \mathbf{x}^*_t = \underbrace{a(t)\mathbf{x}^*_{t-1}}_\textrm{content shift} + \underbrace{b(t){\boldsymbol{\delta}}}_\textrm{trigger shift} + \underbrace{c(t)\boldsymbol{\epsilon}}_\textrm{noise shift},
\end{equation}
where $a(t)$, $b(t)$, and $c(t)$ are the content schedulers, trigger schedulers, and noise schedulers, respectively.
Intuitively, the image distribution at the current forward step, $\mathbf{x}^*_t$, is a composite of three components: 
\begin{itemize}
    \item \textbf{Content shift}: Retains an $a(t)$-scaled proportion of the image distribution from the previous step.
    \item \textbf{Trigger shift}: Adds a $b(t)$-scaled proportion of the backdoor trigger to the image distribution.
    \item \textbf{Noise shift}: Introduces a $c(t)$-scaled Gaussian noise to the image distribution. 
\end{itemize}

As visualized in Fig.~\ref{figure:distribution-shift}, the ``noise shift" term introduces incremental Gaussian noise at each timestep, gradually deteriorating the image. Simultaneously, the ``trigger shift" term iteratively adjusts the image distribution toward the trigger distribution, culminating in a \textit{noisy trigger} distribution, $\mathbf{x}^*_T\sim \mathcal{N}({\boldsymbol{\delta}},\mathbf{I})$, at the final diffusion timestep.

\subsubsection{Multi-Timestep Distribution-Shift Loss}
As the backdoor forward process gradually incorporates the trigger into the image distribution at each timestep, the backdoor backward process conversely removes a proportional amount of the trigger at the corresponding timestep. This is achieved by training a UNet ${\boldsymbol{\theta}}$ to predict not only the content shift (similar to benign models), but also the trigger shift in its output. 
Since the UNet's output, $\boldsymbol{\epsilon}_{\boldsymbol{\theta}}(\mathbf{x}^*_t,t)$, is accessible, the defenders can potentially invert the backdoor trigger by inferring the trigger shifts from the backward process. Specifically, while estimating the trigger shift, the content shift and noise shift can be treated as negligible noise and ignored. Consequently, the relationship between the UNet's output and the trigger shift can be approximated as $\boldsymbol{\epsilon}_{\boldsymbol{\theta}}(\mathbf{x}^*_t,t) \approx \lambda_t{\boldsymbol{\delta}}$, where $\lambda_t$ is the scale of the trigger shift (referred to as \textit{trigger-shift scale}) at timestep $t$. If $\boldsymbol{\epsilon}_{\boldsymbol{\theta}}(\mathbf{x}^*_t,t)$ and $\lambda_t$ are known for multiple timesteps, the defender can learn the trigger ${\boldsymbol{\delta}}$ by minimizing the following multi-timestep distribution-shift loss function: 
\begin{equation}
\label{equation:loss-mds}
    L_{MDS} = \ex_{\boldsymbol{\epsilon},t}\|\boldsymbol{\epsilon}_{\boldsymbol{\theta}}(\mathbf{x}^*_t({\boldsymbol{\delta}},\boldsymbol{\epsilon}), t) - \lambda_t{\boldsymbol{\delta}} \|^2_2.
\end{equation}

However, only the UNet's output $\boldsymbol{\epsilon}_{\boldsymbol{\theta}}(\mathbf{x}^*_t,t)$ is directly accessible, while the trigger-shift scales ${\boldsymbol{\lambda}} = \{\lambda_0,...,\lambda_t,...,\lambda_T\}$ are intractable due to the black-box nature of neural networks (in this case, the UNet). Consequently, estimating the trigger-shift scales ${\boldsymbol{\lambda}}$ becomes the key challenge in inverting backdoor triggers.
In an existing work named Elijah, An et al.~\cite{an2024elijah} heuristically choose $\lambda_T=0.5$ and use it to invert the trigger. However, their approach only leverages the first backward timestep $T$, leading to suboptimal performance in trigger inversion. 

To address this limitation, our work introduces a novel method for estimating the trigger-shift scales, ${\boldsymbol{\lambda}}$, across multiple timesteps. 
Specifically, since the backward process is essentially the reverse of the forward process, the amount of trigger added during the forward process at timestep $t$ should correspond closely to the amount removed by the UNet during the backward process at the same timestep. In other words, the ``trigger shift" component in the backward transition must align with that of the forward transition. Based on this insight, we analyze the trigger shift behavior in both forward and backward transitions across three state-of-the-art backdoor methods, as detailed below.

\textbf{BadDiffusion.} By reparameterizing the equation (\ref{equation:backdoor-forward-baddiffusion}), the forward transition of BadDiffusion can be expressed as
\begin{equation}
\label{equation:baddiffusion-forward-reparameter}
    \mathbf{x}^*_{t} = \sqrt{\alpha_t}\mathbf{x}^*_{t-1} - (1-\sqrt{\alpha_t}){\boldsymbol{\delta}} + (1-\alpha_t)\boldsymbol{\epsilon}.
\end{equation}
\begin{equation}
\label{equation:baddiffusion-forward-reparameter1}
    \implies \mathbf{x}^*_{t-1} = \frac{\mathbf{x}^*_t}{\sqrt{\alpha_t}} - \underbrace{\frac{1-\sqrt{\alpha_t}}{\sqrt{\alpha_t}}{\boldsymbol{\delta}}}_\textrm{trigger shift} + \frac{1-\alpha_t}{\sqrt{\alpha_t}}\boldsymbol{\epsilon}.
\end{equation}

On the other hand, the backdoor backward sampling of BadDiffusion~\cite{Chou2023CVPR} is given as
\begin{equation}
\label{equation:badduffusion-sampling}
    \mathbf{x}_{t-1} = \frac{\mathbf{x}^*_t}{\sqrt{\alpha_t}} - \underbrace{\frac{1-\alpha_t}{\sqrt{\alpha_t}\sqrt{1-\Bar{\alpha}_t}}\boldsymbol{\epsilon}_{\boldsymbol{\theta}}(\mathbf{x}^*_t,t)}_\textrm{approximated trigger shift} + \sigma_t\boldsymbol{\epsilon}.
\end{equation}

By matching the two trigger shifts in equations (\ref{equation:baddiffusion-forward-reparameter1}) and (\ref{equation:badduffusion-sampling}), we derive
\begin{equation}
    \boldsymbol{\epsilon}_{\boldsymbol{\theta}}(\mathbf{x}^*_t,t) \approx \frac{(1-\sqrt{\alpha_t})\sqrt{1-\Bar{\alpha}_t}}{1-\alpha_t}{\boldsymbol{\delta}} = \lambda_t{\boldsymbol{\delta}},
\end{equation}
which leads to the trigger-shift scale:
\begin{equation}
\label{equation:baddiffusion-trigger-scale}
    \lambda_t \approx \frac{(1-\sqrt{\alpha_t})\sqrt{1-\Bar{\alpha}_t}}{1-\alpha_t}.
\end{equation}

\textbf{TrojDiff.} With a similar analysis as above, we can derive the trigger shift of TrojDiff's backdoor forward transition from the equation (\ref{equation:backdoor-forward-trojdiff}) as $\frac{k_t(1-\gamma)}{\sqrt{\alpha_t}}{\boldsymbol{\delta}}$.

The approximated trigger shift from the backward sampling process can be derived from~\cite{chen2023trojdiff}, which is $\frac{1-\alpha_t}{\sqrt{1-\Bar{\alpha_t}}\sqrt{\alpha_t}}\boldsymbol{\epsilon}_{\boldsymbol{\theta}}(\mathbf{x}^*_t,t)$.
By matching the two trigger shifts, the trigger-shift scale of TrojDiff's denoising process is derived as
\begin{equation}
\label{equation:trojdiff-trigger-scale}
    \lambda_t \approx \frac{k_t(1-\gamma)\sqrt{1-\Bar{\alpha}_t}}{1-\alpha_t}.
\end{equation}

\textbf{VillanDiffusion.} For VillanDiffusion, the trigger shift of the backdoor diffusion transition can be derived from (\ref{equation:backdoor-forward-villandiffusion}), resulting in $\frac{h_t}{v_t}{\boldsymbol{\delta}}$, where $h_t=\Bar{\rho}(t)-\sum_{i=1}^{t-1} \left( \left( \prod_{j=i+1}^{t}k_j \right) h_i \right)$ and $v_t=\frac{\Bar{\alpha}(t)}{\Bar{\alpha}(t-1)}$ \cite{chou2024villandiffusion}.

The approximated trigger shift from the reversed transition is 
$\frac{1-\alpha_t}{\sqrt{\alpha_t}\sqrt{1-\Bar{\alpha}_t}}\boldsymbol{\epsilon}_{\boldsymbol{\theta}}(\mathbf{x}^*_t,t)$.

As a result, the trigger-shift scale of VillanDiffusion is:
\begin{equation}
\label{equation:villandiffusion-trigger-scale}
    \lambda_t \approx  \frac{h_t\sqrt{\alpha(t)}\sqrt{1-\Bar{\alpha}(t)}}{v_t(1-\alpha(t))}.
\end{equation}


In \textbf{white-box} defense settings, where the model's coefficients are accessible, the defender can directly compute the trigger-shift scales ${\boldsymbol{\lambda}}$ for a given backdoor method using the corresponding equations (\ref{equation:baddiffusion-trigger-scale}), (\ref{equation:trojdiff-trigger-scale}), or (\ref{equation:villandiffusion-trigger-scale}). These computed scales can then be applied to invert backdoor triggers using the proposed loss function $L_{MDS}$ in (\ref{equation:loss-mds}).

In \textbf{gray-box} defense settings, however, the defender does not have access to the model's coefficients, making it infeasible to directly compute ${\boldsymbol{\lambda}}$. To address this limitation, we propose a practical method to estimate ${\boldsymbol{\lambda}}$, leveraging the following key observation: \textbf{The trigger-shift scales ${\boldsymbol{\lambda}}$ remain invariant across different backdoor triggers and targets, regardless of their shape and size}. This observation is supported by the derived formulas for $\lambda_t$ in equations (\ref{equation:baddiffusion-trigger-scale}), (\ref{equation:trojdiff-trigger-scale}), and (\ref{equation:villandiffusion-trigger-scale}), which do not depend on the specific trigger or target.
Consequently, if the trigger-shift scales for an arbitrary surrogate trigger $\hat{{\boldsymbol{\delta}}}$ are determined, they can also be used as the trigger-shift scales for the true trigger ${\boldsymbol{\delta}}$. To estimate ${\boldsymbol{\lambda}}$, we introduce the following approach: (1) Choose a surrogate trigger $\hat{{\boldsymbol{\delta}}}$. (2) Simulate a backdoor attack targeting a copy of the suspicious diffusion model using $\hat{{\boldsymbol{\delta}}}$. (3) Solve the minimization problem:
\begin{equation}
\label{equation:trigger-scale-minimize}
    \min_{\lambda_t} \left\| \boldsymbol{\epsilon}_{\boldsymbol{\theta}}(\mathbf{x}^*_t(\hat{{\boldsymbol{\delta}}},\boldsymbol{\epsilon}),t) - \lambda_t\hat{{\boldsymbol{\delta}}}  \right\|^2_2.
\end{equation}

This minimization problem can be efficiently solved by setting the derivative to zero, yielding the optimal $\lambda_t$:
\begin{equation}
\label{equation:trigger-scale-solution}
    \lambda_t = \frac{\boldsymbol{\epsilon}_{\boldsymbol{\theta}}(\mathbf{x}^*_t,t)\cdot\hat{{\boldsymbol{\delta}}}}{\| \hat{{\boldsymbol{\delta}}} \|^2_2}.
\end{equation}

\begin{figure}[!t]
    \centering
    \begin{tikzpicture}
        \begin{axis}[
            width=\columnwidth, 
            height=0.75\columnwidth,
            xlabel={Timestep},
            ylabel={Scale of Trigger Distribution Shift},
            xtick={0,1,2,3,4,5,6,7,8,9,10},
            xticklabels={1000, 900, 800, 700, 600, 500, 400, 300, 200, 100, 0},
            legend pos=south west,
            legend style={
                font=\footnotesize,
                at={(0.11, 0.08)}, 
                anchor=south west
            },
            grid=major,
            tick label style={font=\scriptsize},
            label style={font=\small},
            title style={font=\small},
            ylabel near ticks,
            xlabel near ticks
        ]
            \addplot[color=blue, mark=*] coordinates {
                (0, 0.49137247) (1, 0.49966228) (2, 0.50810415) (3, 0.5173604) 
                (4, 0.5252483) (5, 0.5293968) (6, 0.5316261) (7, 0.5102797)
                (8, 0.4498554) (9, 0.36237544) (10, 0.104029305)
            };
            \addlegendentry{Heart-Hat}
            
            \addplot[color=red, mark=square*] coordinates {
                (0, 0.49621308) (1, 0.5024591) (2, 0.5108069) (3, 0.5197135)
                (4, 0.5235676) (5, 0.5309484) (6, 0.5312713) (7, 0.49285698)
                (8, 0.42952678) (9, 0.32317564) (10, 0.08059246)
            };
            \addlegendentry{Box18-Cat}
            
            \addplot[color=green, mark=triangle*] coordinates {
                (0, 0.5135837) (1, 0.49893948) (2, 0.4973149) (3, 0.5021725)
                (4, 0.50591534) (5, 0.50766885) (6, 0.5036138) (7, 0.45744997)
                (8, 0.38698292) (9, 0.26725516) (10, 0.05678188)
            };
            \addlegendentry{House-Cat}
            
            \addplot[color=orange, mark=o] coordinates {
                (0, 0.49999624) (1, 0.5007695) (2, 0.505613) (3, 0.51492083)
                (4, 0.52341) (5, 0.5234597) (6, 0.5137621) (7, 0.48130283)
                (8, 0.41852883) (9, 0.32511112) (10, 0.081482925)
            };
            \addlegendentry{Tree-Hat}
            
            \addplot[color=purple, mark=diamond*] coordinates {
                (0, 0.4965449) (1, 0.5014932) (2, 0.5116682) (3, 0.52264875)
                (4, 0.5329745) (5, 0.5411894) (6, 0.53332937) (7, 0.49325097)
                (8, 0.4037029) (9, 0.31831506) (10, 0.11439891)
            };
            \addlegendentry{Warning1-Cat}
            
            \addplot[color=cyan, mark=star] coordinates {
                (0, 0.5099528) (1, 0.498611) (2, 0.5008345) (3, 0.5094778)
                (4, 0.5174225) (5, 0.5176077) (6, 0.5143738) (7, 0.48389056)
                (8, 0.42565283) (9, 0.33036435) (10, 0.07818606)
            };
            \addlegendentry{House-Hat}
            
            \addplot[color=magenta, mark=pentagon*] coordinates {
                (0, 0.5060673) (1, 0.5020121) (2, 0.50401765) (3, 0.50721186)
                (4, 0.5102461) (5, 0.5063292) (6, 0.48909774) (7, 0.44982737)
                (8, 0.38937548) (9, 0.28901434) (10, 0.06240766)
            };
            \addlegendentry{Tree-Cat}
            
            \addplot[color=black, mark=hexagon*] coordinates {
                (0, 0.4888207) (1, 0.50170094) (2, 0.51581764) (3, 0.52808595)
                (4, 0.5390451) (5, 0.5449231) (6, 0.53765345) (7, 0.5069194)
                (8, 0.45654058) (9, 0.36222264) (10, 0.10839939)
            };
            \addlegendentry{Warning2-Cat}
        \end{axis}
    \end{tikzpicture}
    \caption{The scale of trigger shift over all denoising timesteps for different trigger-target pairs.}
    \label{fig:trigger_shift}
\end{figure}
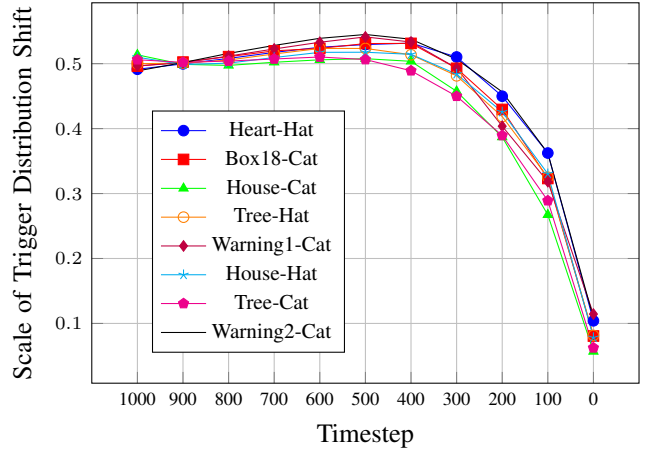

Once ${\boldsymbol{\lambda}} = \{\lambda_0,...,\lambda_t,...,\lambda_T\}$ is obtained, it can be utilized to invert backdoor triggers via the loss function $L_{MDS}$ in the equation (\ref{equation:loss-mds}).
Fig.~\ref{fig:trigger_shift} illustrates the trigger-shift scales ${\boldsymbol{\lambda}}$ computed using our proposed method with different triggers and targets. The results show that ${\boldsymbol{\lambda}}$ only varies slightly across different trigger-target pairs due to the stochastic nature of diffusion models, but its overall form remains consistent. This consistency empirically supports our observation regarding the invariance of ${\boldsymbol{\lambda}}$. This experiment also explains why Elijah\cite{an2024elijah} can perform properly by heuristically choosing $\lambda_T=0.5$ for the first denoising step.

\begin{figure*}[h!]
	\centering
	\includegraphics[scale=0.12]{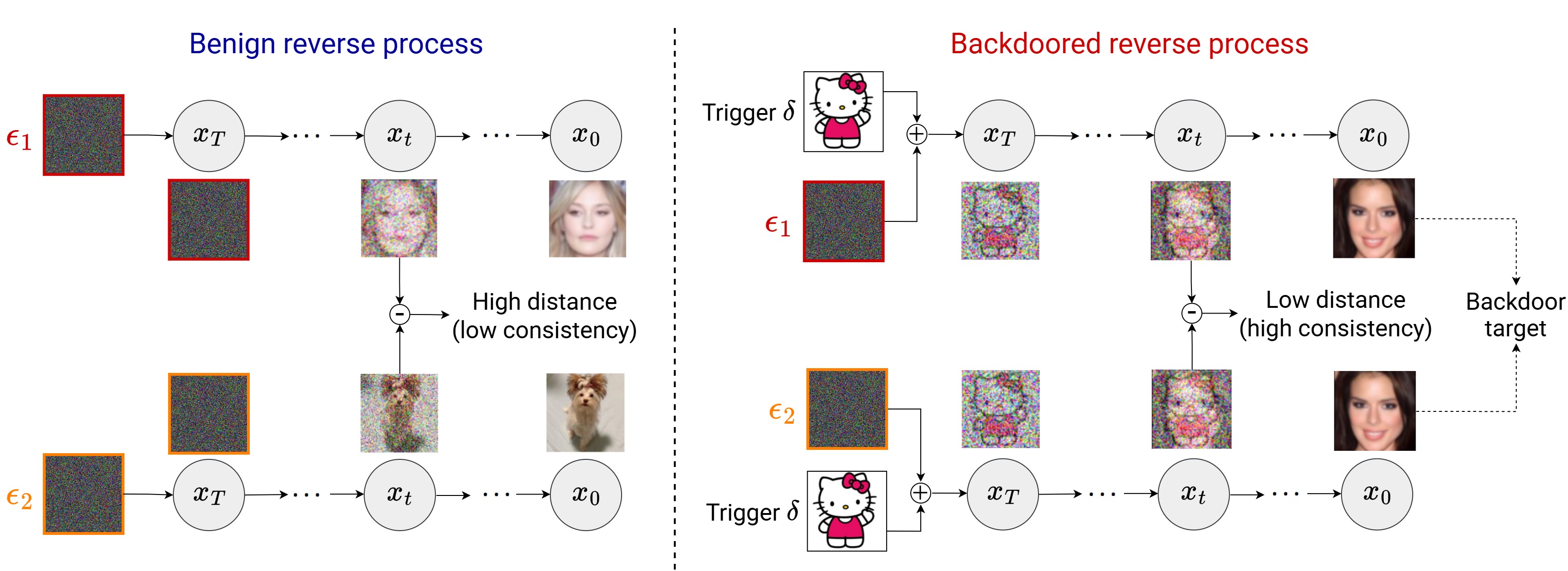}
	\caption{An illustration for the ``consistency effect" caused by the backdoored reverse process when sampling with different input noises.}
	\label{figure:denoising-consistency}
\end{figure*}

\subsubsection{Denoising-Consistency Loss}
While the above loss function $L_{MDS}$ investigates the trigger shift in diffusion models, the content shift also reveals critical traces of backdoor attacks. As described in Fig.~\ref{figure:denoising-consistency}, the backward sampling process of a backdoored diffusion model produces diverse generation results for different input noises. However, when the trigger is stamped in the input noise, the generation results become highly consistent across different generation attempts, as the model consistently produces the backdoor target. This phenomenon highlights the consistency of content shifts in the presence of the backdoor trigger, whereas they remain inconsistent when the noise varies without the trigger.

While estimating the content shift, the trigger shift can be considered as negligible noise and ignored. 
Thus, to quantify this consistency, the content shift can be approximated by subtracting the noise shift from the UNet's output $(\boldsymbol{\epsilon}_{\boldsymbol{\theta}}(\mathbf{x}^*_t({\boldsymbol{\delta}},\boldsymbol{\epsilon}),t) - n(\boldsymbol{\epsilon}))$, where $n(\boldsymbol{\epsilon})$ represents the noise shift, which is known and depends on the specific type of diffusion model.
To invert backdoor triggers, we leverage the observation that the content shifts caused by two different input noises, $\boldsymbol{\epsilon}_1$ and $\boldsymbol{\epsilon}_2$, must be consistent in the presence of a trigger. Based on this principle, we propose to learn $\boldsymbol{\delta}$ by minimizing the following denoising-consistency loss function:
\begin{equation}
\label{equation:loss-dc}
\begin{split}
    L_{DC} = \ex_{\boldsymbol{\epsilon}_1, \boldsymbol{\epsilon}_2} \|(\boldsymbol{\epsilon}_{\boldsymbol{\theta}}(\mathbf{x}^*_t({\boldsymbol{\delta}},\boldsymbol{\epsilon}_1),t) - n(\boldsymbol{\epsilon}_1)) \\ - (\boldsymbol{\epsilon}_{\boldsymbol{\theta}}(\mathbf{x}^*_t({\boldsymbol{\delta}},\boldsymbol{\epsilon}_2),t) - n(\boldsymbol{\epsilon}_2))\|^2_2
\end{split}.
\end{equation}

This loss function requires only the UNet's output and does not rely on additional coefficients, making it applicable in both white-box and gray-box defense settings.

\begin{algorithm}
\small
\caption{Trigger Inversion}\label{algorithm:trigger-inversion}
\textbf{Input:} Suspicious diffusion model ${\boldsymbol{\theta}}$; number of iterations, $n_{MDS}$ and $n_{DC}$, for the two losses  \\
\textbf{Output:} Inverted trigger ${\boldsymbol{\delta}}$

\begin{algorithmic}[1]
\IF{Gray-Box Defense}
\STATE // \textit{Estimate the trigger-shift scales}
\STATE Select a surrogate trigger $\hat{{\boldsymbol{\delta}}}$.
\STATE Use $\hat{{\boldsymbol{\delta}}}$ to simulate a backdoor attack on the model ${\boldsymbol{\theta}}$.
\FOR{$t = T, T-1,...,1$} 
    \STATE Predict $\boldsymbol{\epsilon}_{\boldsymbol{\theta}}(t)$ with $\hat{{\boldsymbol{\delta}}}$ stamped to the UNet's input noise.
    \STATE $\lambda_t = (\boldsymbol{\epsilon}_{\boldsymbol{\theta}}(t)\cdot\hat{{\boldsymbol{\delta}}}) / || \hat{{\boldsymbol{\delta}}} ||^2_2$.
\ENDFOR
\ENDIF

\STATE // \textit{Trigger Inversion}
\STATE Initialize a learnable trigger ${\boldsymbol{\delta}}$, starting from Gaussian noise.

\STATE // \textit{Invert trigger using $L_{MDS}$}
\FOR{$epoch = 0, 1,...,n_{MDS}$} 
    \STATE Sample a random timestep $t \in [0, T]$.
    \STATE Predict $\boldsymbol{\epsilon}_{\boldsymbol{\theta}}(t)$ with ${\boldsymbol{\delta}}$ stamped to the UNet's input noise.
    \STATE Compute $L_{MDS}$ based on (\ref{equation:loss-mds}).
    \STATE Backprop to optimize ${\boldsymbol{\delta}}$.
\ENDFOR

\STATE // \textit{Invert trigger using $L_{DC}$}
\FOR{$epoch = 0, 1,...,n_{DC}$} 
    \STATE Sample a random timestep $t \in [0, T]$.
    \STATE Predict $\boldsymbol{\epsilon}_{\boldsymbol{\theta}}(t)$ with ${\boldsymbol{\delta}}$ stamped to the UNet's input noise.
    \STATE Compute $L_{MDS}$ based on (\ref{equation:loss-dc}).
    \STATE Backprop to optimize ${\boldsymbol{\delta}}$.
\ENDFOR

\RETURN ${\boldsymbol{\delta}}$
\end{algorithmic}
\end{algorithm}

\subsubsection{Learning Strategy}
In practice, we observed that the loss function $L_{MDS}$ tends to produce inverted triggers with higher ASR, whereas $L_{DC}$ generates triggers that more closely resemble the ground-truth trigger. To leverage the strengths of both approaches, we adopt a sequential strategy instead of combining the two losses into a single loss function. Specifically, we first use $L_{MDS}$ to invert the backdoor trigger, optimizing for higher ASR. Subsequently, we refine this inverted trigger using $L_{DC}$, enhancing its resemblance to the ground-truth trigger. This sequential approach consistently achieves the highest trigger inversion performance in our experiments.

To compute $L_{MDS}$ and $L_{DC}$, we need the UNet's output, $\boldsymbol{\epsilon}_{\boldsymbol{\theta}}(\mathbf{x}^*_t(\hat{{\boldsymbol{\delta}}},\boldsymbol{\epsilon}),t)$, at a specific timestep $t$. This is obtained by running the backward process for $T-t$ steps, starting from the first denoising step $T$. In this process, the UNet's output from each denoising step serves as the input for the next step. Consequently, gradient descent for trigger inversion involves backpropagation across $T-t$ backward steps, creating a long computational graph. Due to the high computational demands of this approach, we limit the process to the first $N$ denoising steps, where we set $N = 50$ in our experimental studies. In each iteration, we randomly sample a backward step within 0 and $N$. This range balances computational feasibility with inversion performance, as earlier denoising steps carry more significant backdoor information than later steps (as shown in Fig.~\ref{fig:trigger_shift}). Additionally, we use a small batch size to manage memory usage and ensure stable optimization during gradient descent.

\subsection{Backdoor Detection}\label{subsection:backdoor-detection}
Although our trigger inversion method can efficiently invert triggers from backdoored diffusion models, it cannot directly determine whether a suspicious model is backdoored. To address this limitation, we propose a hybrid backdoor detection technique that combines trigger-based detection and generation-based detection, as detailed below. It should be noted that the backdoor detection method presented in this section is conducted after the trigger inversion stage, meaning that we already obtained a candidate trigger ${\boldsymbol{\delta}}$ inverted by the proposed trigger inversion method.

\subsubsection{Generation-Based Detection}
This approach evaluates the samples generated by the suspicious diffusion model, both with and without the presence of the inverted trigger, to determine whether a backdoor attack exists. Specifically, if the model is indeed backdoored, embedding the inverted trigger into the model's input noise should consistently yield the backdoor target across different random noise inputs. Conversely, if the model is benign, the inverted trigger will produce diverse samples across multiple generation attempts with varying input noises. 
To quantify this behavior, we generate $K$ samples using the inverted trigger and calculate the similarity between every pair of generated samples. This results in a similarity score (SIM) for the suspicious model ${\boldsymbol{\theta}}$:
\begin{equation}
\label{equation:similarity-score}
    S({\boldsymbol{\theta}},{\boldsymbol{\delta}}) = \ex_{i,j \in [1, K], i \neq j} \| F({\boldsymbol{\theta}}, {\boldsymbol{\delta}},\boldsymbol{\epsilon}_i) - F({\boldsymbol{\theta}},{\boldsymbol{\delta}},\boldsymbol{\epsilon}_j)  \|,
\end{equation}
where ${\boldsymbol{\theta}}$ represents the suspicious diffusion model, $F(\cdot)$ denotes the sampling function that takes the model ${\boldsymbol{\theta}}$, the inverted trigger ${\boldsymbol{\delta}}$, and the input noise $\boldsymbol{\epsilon}$ as inputs, then produces the image sample. Intuitively, a higher SIM score indicates that the diffusion model generates more diverse samples.

Then, we compute the SIM score in case there is no trigger stamped in the input noise. In this case, we simply replace ${\boldsymbol{\delta}}$ in the equation ($\ref{equation:similarity-score}$) with Gaussian noise, resulting in $S({\boldsymbol{\theta}},\boldsymbol{\epsilon})$. Essentially, $S({\boldsymbol{\theta}},\boldsymbol{\epsilon})$ should be significantly higher than $S({\boldsymbol{\theta}},{\boldsymbol{\delta}})$ if ${\boldsymbol{\theta}}$ is a backdoor model, and vice versa. In our work, we determine a model as a potential backdoor model if $S({\boldsymbol{\theta}},\boldsymbol{\epsilon}) \geq k \cdot S({\boldsymbol{\theta}},{\boldsymbol{\delta}})$, with $k=5$ in our experiments.

\subsubsection{Trigger-Based Detection}
In real-world attacks, it is shown that the attacker can backdoor a diffusion model with multiple backdoor targets~\cite{chen2023trojdiff}. In such cases, the similarity score $S({\boldsymbol{\theta}},{\boldsymbol{\delta}})$ of a backdoored model may still be high, as the diverse backdoor targets could deceive the generation-based detection method. To address this problem, we employ a second approach that analyzes the inverted trigger itself rather than the generated samples. This approach builds on the observation made in~\cite{mo2024terd}, in which inverting trigger of benign models will eventually converge to a nearly full-zero solution. Based on this, we compute the Kullback-Leibler (KL) divergence~\cite{kullback1951information} between the distribution of the inverted trigger and the benign Gaussian distribution. Intuitively, an abnormally high KL divergence suggests the presence of a backdoor. In practice, we follow the work in~\cite{mo2024terd} to compute the divergence across multiple benign diffusion models to establish a threshold for backdoor detection. If the KL divergence of the suspicious model exceeds this threshold, it is considered as backdoored.

However, this approach can be inefficient if the attacker intentionally selects triggers that closely resemble Gaussian noise, or applies techniques to make the trigger indistinguishable from Gaussian noise~\cite{lilearnable}. Therefore, we apply both generation-based and trigger-based detection methods, leveraging their complementary strengths while tackling the weaknesses of each approach. Consequently, a diffusion model is classified as backdoored if at least one of the two methods detects it as a backdoor model.

\begin{figure}[h!]
	\centering
	\includegraphics[scale=0.13]{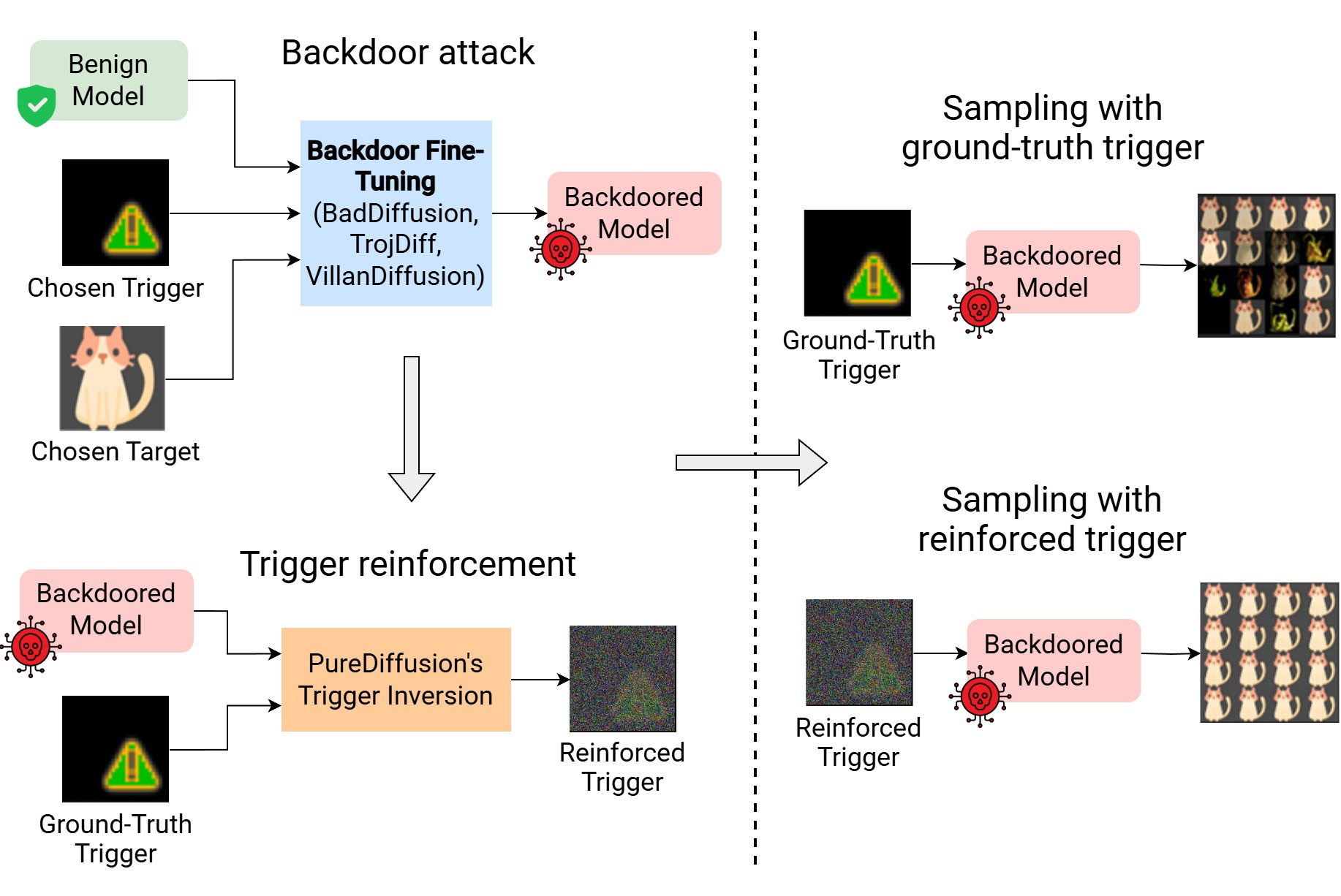}
	\caption{Using our trigger inversion method for backdoor amplification, thereby improving backdoor attack success rate.}
	\label{figure:trigger-reinforcement}
\end{figure}

\subsection{Backdoor Amplification}\label{subsection:backdoor-amplification}
During the evaluation of our defense framework, we observed an interesting phenomenon that suggests an additional potential application of PureDiffusion for backdoor attacks. Specifically, in nearly 50\% of the experimented scenarios, the triggers inverted by PureDiffusion achieved higher ASR than the original ground-truth triggers.
This observation introduces a compelling idea: after backdooring a diffusion model, the original trigger could be further optimized to enhance its attacking performance. By leveraging this optimization, attackers could potentially refine the trigger to achieve superior results, making the backdoor attack even more effective and harder to defend against.

Therefore, although PureDiffusion is primarily designed as a backdoor defense mechanism, its trigger inversion technique can also be adapted into an adversarial attack method. This adaptation, referred to as backdoor amplification, significantly enhances the effectiveness of existing backdoor attacks. As depicted in Fig.~\ref{figure:trigger-reinforcement}, the workflow for backdoor amplification is as follows: First, the attacker chooses a backdoor trigger ${\boldsymbol{\delta}}$ (e.g., the warning sign) and a backdoor target $\mathbf{x}^*_T$ (e.g., the cat image). Second, using exisiting backdoor techniques such as BadDiffusion, TrojDiff, or VillanDiffusion, the attacker backdoors a benign diffusion model, embedding the designated trigger ${\boldsymbol{\delta}}$ into the model. Third, instead of starting the learning process from Gaussian noise, the attacker uses the original ${\boldsymbol{\delta}}$ as the initial input for PureDiffusion's trigger inversion process, employing both the $L_{MDS}$ loss. This results in a \textit{reinforced trigger} ${\boldsymbol{\delta}}^r$, which: (i) Exhibits significantly higher ASR than the original trigger, often approaching 100\% in most cases, and (ii) appears visually distorted after adversarial optimization, making it more challenging for human inspection to detect. The reinforced trigger ${\boldsymbol{\delta}}^r$ can then replace the original one to directly activate the backdoor effect on backdoored models.

From another perspective, our backdoor amplification method also provides a means to reduce the computational cost of backdoor training. Training backdoored models on large datasets is resource-intensive; however, our method enables attackers to reduce the number of training epochs by 10-20 times. By optimizing the trigger with our reinforcement technique, attackers can achieve comparable or even superior attack performance with significantly less training effort.

\begin{algorithm}
\small
\caption{Backdoor Amplification}\label{algorithm:backdoor-amplification}
\textbf{Input:} Benign model ${\boldsymbol{\theta}}$; backdoor trigger ${\boldsymbol{\delta}}$; backdoor target $\mathbf{x}^*_T$; backdoor method $\mathcal{B}(\cdot)$; number of iterations $n_{MDS}$  \\
\textbf{Output:} Backdoored model ${\boldsymbol{\theta}}^*$, reinforced trigger ${\boldsymbol{\delta}}^r$

\begin{algorithmic}[1]
\STATE Backdoor the benign model: ${\boldsymbol{\theta}}^* \leftarrow \mathcal{B}({\boldsymbol{\theta}},{\boldsymbol{\delta}},\mathbf{x}^*_T)$. 
\STATE Initialize the reinforced trigger: ${\boldsymbol{\delta}}^r \coloneq {\boldsymbol{\delta}}$.
\FOR{$epoch = 0, 1,...,n_{MDS}$} 
    \STATE Sample a random timestep $t \in [0, 50]$.
    \STATE Predict $\boldsymbol{\epsilon}_{{\boldsymbol{\theta}}^*}(t)$ using ${\boldsymbol{\delta}}^r$ and ${\boldsymbol{\theta}}^*$.
    \STATE $Loss = \ex_{\boldsymbol{\epsilon}} \left\| \boldsymbol{\epsilon}_{{\boldsymbol{\theta}}^*}(t) - \lambda_t{\boldsymbol{\delta}}  \right\|^2_2$.
    \STATE Backprop the computed loss to optimize ${\boldsymbol{\delta}}^r$.
\ENDFOR
\RETURN ${\boldsymbol{\theta}}^*$, ${\boldsymbol{\delta}}^r$
\end{algorithmic}
\end{algorithm}

\section{Experiments}\label{section:experiment}

\begin{table*}[]
\centering
\caption{Backdoor Detection performance of PureDiffusion compared to existing methods.}
\label{table:backdoor-detection-performance}
\begin{tabular}{@{}llcccccccc@{}}
\toprule
\multicolumn{1}{c}{\multirow{2}{*}{\begin{tabular}[c]{@{}c@{}}Attack \\ Method\end{tabular}}} & \multicolumn{1}{c}{\multirow{2}{*}{\begin{tabular}[c]{@{}c@{}}Defense\\ Method\end{tabular}}} & \multicolumn{4}{c}{Default Triggers} & \multicolumn{4}{c}{Difficult Triggers} \\ \cmidrule(l){3-10} 
\multicolumn{1}{c}{} & \multicolumn{1}{c}{} & ACC $\uparrow$ & TPR $\uparrow$ & TNR $\uparrow$ & \multicolumn{1}{l}{L2D}$\downarrow$ & ACC $\uparrow$ & TPR $\uparrow$ & TNR $\uparrow$ & \multicolumn{1}{l}{L2D}$\downarrow$ \\ \midrule
\multirow{3}{*}{BadDiffusion} & Elijah & 75.01 & 89.12 & 46.77 & 34.61 & 33.40 & 25.04 & 50.11 & 37.56 \\
 & TERD & 97.47 & 96.20 & 100.00 & 18.32 & 41.16 & 11.74 & 100.00 & 28.53 \\
 & Ours & \textbf{100.00} & \textbf{100.00} & \textbf{100.00} & \textbf{16.80} & \textbf{95.09} & \textbf{92.64} & \textbf{100.00} & \textbf{17.47} \\ \midrule
\multirow{3}{*}{TrojDiff} & Elijah & 57.89 & 43.22 & 87.23 & 24.75 & 38.17 & 15.87 & 82.77 & 25.01 \\
 & TERD & 95.03 & 92.54 & 100.00 & \textbf{4.88} & 76.97 & 65.46 & 100.00 & 15.55 \\
 & Ours & \textbf{100.00} & \textbf{100.00} & \textbf{100.00} & 6.02 & \textbf{98.83} & \textbf{98.25} & \textbf{100.00} & \textbf{6.66} \\ \midrule
\multirow{3}{*}{VillanDiffusion} & Elijah & 27.71 & 11.53 & 60.08 & 39.64 & 28.31 & 8.24 & 68.45 & 31.63 \\
 & TERD & 91.26 & 88.85 & 96.09 & 27.50 & 32.58 & 5.75 & 86.23 & 26.77 \\
 & Ours & \textbf{100.00} & \textbf{100.00} & \textbf{100.00} & \textbf{25.54} & \textbf{82.71} & \textbf{76.20} & \textbf{95.74} & \textbf{17.74} \\ \bottomrule
\end{tabular}
\end{table*}

\subsection{Experimental Setup}

\subsubsection{Dataset and Attack Settings}
We evaluate our defense mechanism against state-of-the-art backdoor attacks, including BadDiffusion, TrojDiff, and VillanDiffusion. For DDPMs, we employ all three backdoor methods, whereas score-based models are attacked exclusively using VillanDiffusion. Our experiments primarily use the CIFAR-10 dataset~\cite{krizhevsky2009learning}, which comprises 60,000 images spanning 10 classes, each with a resolution of $32 \times 32$.
To challenge existing defense techniques, we design experiments with more complex triggers, as shown in Fig.\ref{figure:backdoor-settings}, featuring 8 triggers and 2 targets. Trigger inversion and backdoor detection are tested on a total of 100 benign models and 200 backdoored models. The benign diffusion models are sourced from Hugging Face, while the backdoored models are trained using the official codebases provided by the respective authors\cite{Chou2023CVPR, an2024elijah, chou2024villandiffusion}. For fairness, we maintain the same number of backdoor training epochs as specified in their original attack configurations. For evaluating backdoor amplification, we adjust the number of backdoor training epochs from 5 to 100 epochs, experimenting on all 16 possible trigger-target pairs.

\subsubsection{Evaluation Metrics}
Trigger inversion is evaluated on three different metrics that cover both the trigger space and the generation space: (1) \textbf{ASR} measures the effectiveness of the trigger inverted by PureDiffusion in causing the backdoored model to generate the target. Higher ASR indicates better performance. (2) \textbf{SIM} score evaluates the similarity between the generated samples when the inverted trigger is stamped in the model's input. If the trigger is inverted successfully, it must make the model consistently produces the backdoor target, leading to a low distance between the generated samples. Since this scored is based on L1 distance, it is the lower the better. (3) \textbf{L2D} is the L2 distance between the inverted trigger and the ground-truth trigger used to backdoor the model. Lower values denote higher fidelity in trigger inversion.

On the other hand, backdoor detection is evaluated on the following metrics: (1) \textbf{Accuracy (ACC)} assesses PureDiffusion's ability to classify models as either benign or backdoored correctly (2) \textbf{True Positive Rate (TPR)} indicates the proportion of backdoored models that are correctly identified. (4) \textbf{True Negative Rate (TNR)} indicates the proportion of benign models accurately classified as benign. All these metrics are the higher the better.

Backdoor amplification is primarily evaluated using ASR, which measures the efficiency of the reinforced trigger compared to the original ground-truth trigger.

\subsubsection{Baseline Defense Methods}
Before PureDiffusion, only Elijah~\cite{an2024elijah} and TERD~\cite{mo2024terd} were designed to counter backdoor attacks with both trigger inversion and backdoor detection capabilities. We utilize these two frameworks as baselines for evaluating our method, maintaining the same hyperparameter settings as specified in their original papers and source code.settings with the original papers and their provided source code.

\subsubsection{PureDiffusion Settings}
Our trigger inversion is split into two stages. In the first stage, the trigger is inverted using the $L_{MDS}$ loss over 30 epochs, with a batch size of 8 and a learning rate of 0.1. Then, the inverted trigger is further refined using the $L_{DC}$ loss over 500 epochs, with a batch size of 16 and a learning rate of 0.5. For backdoor detection, we evaluate PureDiffusion on 300 diffusion models, comprising 100 benign models and 200 backdoored models, utilizing the triggers inverted by our proposed method. For backdoor amplification, the ground-truth triggers are reinforced using a setup similar to the trigger inversion process. However, the number of training epochs for $L_{DC}$ in this case is reduced to 100.

\begin{table*}[]
\centering
\caption{Performance of trigger inversion compared to existing methods, experimented on difficult trigger-target pairs.}
\label{table:trigger-inversion-performance}
\begin{tabular}{@{}lclccc|lclccc@{}}
\toprule
\multicolumn{6}{c|}{DDPM-Based Models} & \multicolumn{6}{c}{Score-Based Models} \\ \midrule
\multicolumn{1}{c}{Trigger} & Target & \multicolumn{1}{c}{Method} & SIM$\downarrow$ & ASR $\uparrow$ & L2D$\downarrow$ & \multicolumn{1}{c}{Trigger} & Target & \multicolumn{1}{c}{Method} & SIM $\downarrow$ & ASR $\uparrow$ & L2D $\downarrow$ \\ \midrule
\multirow{3}{*}{Stop 14} & \multirow{3}{*}{Hat} & Elijah & 0.39487 & 28.42 & 38.72 & \multirow{3}{*}{Stop 14} & \multirow{3}{*}{Hat} & Elijah & 0.21465 & 41.93 & 36.75 \\
 &  & TERD & X & 0 & 40.32 &  &  & TERD & 0.01145 & 67.52 & 25.12 \\
 &  & Ours & \textbf{0.00061} & \textbf{91.32} & \textbf{15.63} &  &  & Ours & \textbf{0.00045} & \textbf{99.65} & \textbf{13.15} \\ \midrule
\multirow{3}{*}{Stop 18} & \multirow{3}{*}{Hat} & Elijah & X & 0 & 40.37 & \multirow{3}{*}{Stop 18} & \multirow{3}{*}{Hat} & Elijah & 0.21250 & 56.35 & 35.01 \\
 &  & TERD & X & 0 & 39.89 &  &  & TERD & 0.26787 & 45.77 & 25.78 \\
 &  & Ours & \textbf{0.00082} & \textbf{69.35} & \textbf{26.26} &  &  & Ours & \textbf{0.00042} & \textbf{99.12} & \textbf{12.21} \\ \midrule
\multirow{3}{*}{House} & \multirow{3}{*}{Cat} & Elijah & 0.00081 & 91.23 & 29.33 & \multirow{3}{*}{House} & \multirow{3}{*}{Cat} & Elijah & 0.00087 & 90.11 & 18.02 \\
 &  & TERD & 0.00078 & 93.67 & 13.89 &  &  & TERD & 0.00076 & 96.75 & 16.42 \\
 &  & Ours & \textbf{0.00041} & \textbf{99.92} & \textbf{12.43} &  &  & Ours & \textbf{0.00040} & \textbf{98.67} & \textbf{13.68} \\ \midrule
\multirow{3}{*}{Tree} & \multirow{3}{*}{Hat} & Elijah & 0.36782 & 50.72 & 35.21 & \multirow{3}{*}{Tree} & \multirow{3}{*}{Hat} & Elijah & 0.26544 & 58.77 & 37.65 \\
 &  & TERD & X & 0 & 41.23 &  &  & TERD & 0.30012 & 46.94 & 28.32 \\
 &  & Ours & \textbf{0.00072} & \textbf{88.76} & \textbf{23.26} &  &  & Ours & \textbf{0.00056} & \textbf{98.18} & \textbf{14.91} \\ \midrule
\multirow{3}{*}{Warning 1} & \multirow{3}{*}{Hat} & Elijah & 0.00125 & 78.43 & 31.11 & \multirow{3}{*}{Warning 1} & \multirow{3}{*}{Hat} & Elijah & 0.14426 & 72.58 & 33.69 \\
 &  & TERD & X & 0 & 39.23 &  &  & TERD & 0.43256 & 8.66 & 36.42 \\
 &  & Ours & \textbf{0.00065} & \textbf{89.23} & \textbf{14.32} &  &  & Ours & \textbf{0.00079} & \textbf{92.54} & \textbf{18.44} \\ \midrule
\multirow{3}{*}{House} & \multirow{3}{*}{Hat} & Elijah & 0.00068 & 92.47 & 31.98 & \multirow{3}{*}{House} & \multirow{3}{*}{Hat} & Elijah & 0.02478 & 88.86 & 28.33 \\
 &  & TERD & 0.00057 & 96.44 & 12.84 &  &  & TERD & 0.01465 & 92.92 & 20.01 \\
 &  & Ours & \textbf{0.00040} & \textbf{98.78} & \textbf{12.72} &  &  & Ours & \textbf{0.00062} & \textbf{99.34} & \textbf{19.88} \\ \midrule
\multirow{3}{*}{Warning 1} & \multirow{3}{*}{Cat} & Elijah & 0.00062 & 95.45 & 30.32 & \multirow{3}{*}{Warning 1} & \multirow{3}{*}{Cat} & Elijah & 0.16894 & 56.81 & 32.19 \\
 &  & TERD & 0.00055 & 97.21 & 13.01 &  &  & TERD & 0.20011 & 43.03 & 28.00 \\
 &  & Ours & \textbf{0.00040} & \textbf{99.23} & \textbf{12.80} &  &  & Ours & \textbf{0.00997} & \textbf{77.79} & \textbf{20.02} \\ \midrule
\multirow{3}{*}{Warning 2} & \multirow{3}{*}{Cat} & Elijah & 0.56724 & 12.04 & 36.44 & \multirow{3}{*}{Warning 2} & \multirow{3}{*}{Cat} & Elijah & X & 0 & 42.22 \\
 &  & TERD & X & 0 & 41.21 &  &  & TERD & X & 0 & 37.86 \\
 &  & Ours & \textbf{0.00089} & \textbf{78.89} & \textbf{22.56} &  &  & Ours & \textbf{0.19895} & \textbf{56.23} & \textbf{22.68} \\ \bottomrule
\end{tabular}
\end{table*}

\subsection{Defense Performance}
To evaluate the defense performance of PureDiffusion against backdoor attacks, we experiment it alongside Elijah and TERD using two sets of backdoor triggers. The first set includes default triggers provided by the three backdoor methods, such as the white/gray box and the small stop sign. The second set contains our newly chosen triggers that are more challenging for trigger inversion, such as the warning signs and the big stop sign.
The evaluation results are summarized in Table~\ref{table:backdoor-detection-performance}. 

PureDiffusion outperformed both baseline methods, achieving 100\% accuracy, TPR, and TNR across all cases. Additionally, it demonstrated the best visual fidelity, with inverted triggers showing the lowest L2 distance (L2D) from the ground-truth triggers in most scenarios. While TERD also performed relatively well on default triggers, Elijah’s performance was notably lower across all metrics.

However, experiments on difficult triggers reveal the inefficiency of both Elijah and TERD in both trigger inversion and backdoor detection, as shown in Table~\ref{table:backdoor-detection-performance}. Specifically, while PureDiffusion maintained robust performance, with only a slight decrease in detection accuracy and inversion quality under the difficult trigger settings, the performance of Elijah and TERD dropped dramatically in all four metrics. Notably, Elijah and TERD struggled to invert the triggers effectively, leading to high L2D values. Consequently, their detection methods often misclassified models as negative, resulting in high TNR but low TPR and accuracy.

\subsection{Evaluation on Extended Trigger-Target Pairs}
Additional experiments on challenging trigger patterns further highlight the performance gap between PureDiffusion and the baseline methods. As detailed in Table~\ref{table:trigger-inversion-performance}, PureDiffusion's trigger inversion consistently outperforms both Elijah and TERD across all metrics. 

Regarding performance in the trigger space, triggers inverted by PureDiffusion exhibit high fidelity, with significantly lower L2 distances from the ground-truth triggers compared to those produced by the baseline methods. This demonstrates PureDiffusion's superior ability to accurately reconstruct the backdoor triggers.

In terms of performance in the generation space, triggers inverted by PureDiffusion achieve the highest backdoor ASR, effectively causing the backdoored models to generate the intended backdoor targets. The generated samples are highly consistent and accurately resemble the backdoor targets (as shown by the SIM score), further validating the quality of the inverted triggers.
Conversely, triggers inverted by Elijah and TERD are often of low quality, leading to poor attacking performance in both SIM score and ASR. In some cases, such as when the trigger is the large stop sign (i.e., STOP-18) and the target is the pink hat, both baselines fail entirely, resulting in 0\% ASR and extremely high L2 distances. For cases where the baselines cannot invert the trigger, the resulting generation outputs are highly random. Consequently, the SIM scores are not meaningful in these scenarios and are marked as ``X" in Table~\ref{table:trigger-inversion-performance}.

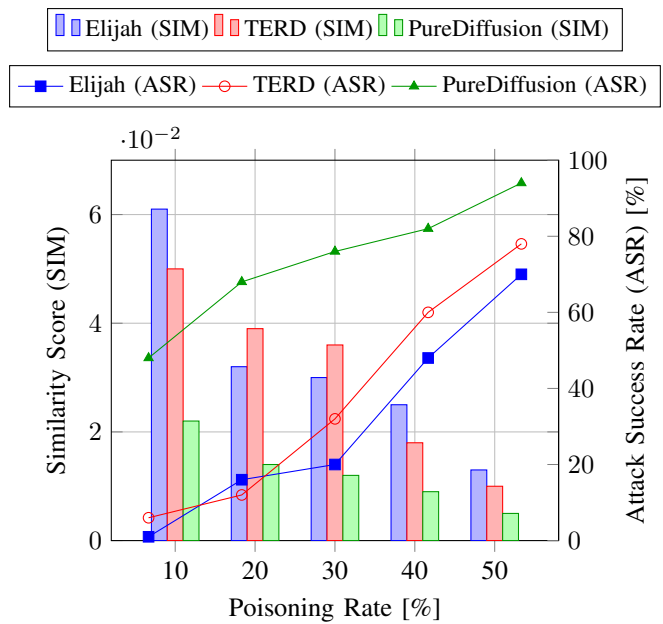
\begin{figure}[!t]
\centering
\begin{tikzpicture}

\begin{axis}[
    width=0.85\columnwidth,
    height=0.75\columnwidth,
    xlabel={Poisoning Rate [\%]},
    ylabel={Similarity Score (SIM)},
    ylabel style={yshift=0pt},
    ybar=0pt,
    bar width=6pt,
    symbolic x coords={10, 20, 30, 40, 50},
    xtick=data,
    ymin=0, ymax=0.07,
    enlarge x limits=0.2,
    grid=major,
    legend style={at={(0.5,1.4)}, anchor=north, legend columns=3, font=\small}
]

\addplot[fill=blue!30, draw=blue] coordinates {(10,0.061) (20,0.032) (30,0.030) (40,0.025) (50,0.013)};
\addlegendentry{Elijah (SIM)}

\addplot[fill=red!30, draw=red] coordinates {(10,0.050) (20,0.039) (30,0.036) (40,0.018) (50,0.010)};
\addlegendentry{TERD (SIM)}

\addplot[fill=green!30, draw=green!60!black] coordinates {(10,0.022) (20,0.014) (30,0.012) (40,0.009) (50,0.005)};
\addlegendentry{PureDiffusion (SIM)}

\end{axis}

\begin{axis}[
    width=0.85\columnwidth,
    height=0.75\columnwidth,
    axis x line=none, 
    axis y line*=right, 
    ylabel={Attack Success Rate (ASR) [\%]},
    ylabel style={yshift=0pt},
    ymin=0, ymax=100,
    ytick={0,20,40,60,80,100},
    xtick=data,
    legend style={at={(0.5,1.25)}, anchor=north, legend columns=3, font=\small}
]

\addplot[color=blue, mark=square*] coordinates {(10,1) (20,16) (30,20) (40,48) (50,70)};
\addlegendentry{Elijah (ASR)}

\addplot[color=red, mark=o] coordinates {(10,6) (20,12) (30,32) (40,60) (50,78)};
\addlegendentry{TERD (ASR)}

\addplot[color=green!60!black, mark=triangle*] coordinates {(10,48) (20,68) (30,76) (40,82) (50,94)};
\addlegendentry{PureDiffusion (ASR)}

\end{axis}

\end{tikzpicture}
\caption{Performance of our trigger inversion method in different backdoor poisoning rates.}
\label{fig:poisoning_rate_comparison}
\end{figure}

\subsection{Evaluation on Varying Poisoning Rate}
Backdooring diffusion models with varying poisoning rates has a significant impact on their attacking performance. A higher poisoning rate generally results in a higher ASR since the models are exposed to more backdoor data, allowing them to recognize the triggers more effectively.
In this section, we analyze the effect of backdoor poisoning rates on defense performance. 

Specifically, we backdoor diffusion models with poisoning rates ranging from 10\% to 50\%, using the small stop sign (STOP-14) as trigger and the pink hat as target. Next, we use PureDiffusion, along with the two baseline methods, to invert the trigger and evaluate its ASR and SIM. As shown in Fig.~\ref{fig:poisoning_rate_comparison}, it is particularly challenging for both TERD and Elijah to invert triggers when the poisoning rate is low (e.g., around 10\%). Their inability to reconstruct triggers in such scenarios leads to poor attacking performance. In contrast, PureDiffusion demonstrates a superior capability to invert triggers effectively even at low poisoning rates, maintaining relatively high ASR and acceptable SIM values.

Overall, the performance of all three frameworks improves as the poisoning rate increases. However, PureDiffusion consistently outperforms the baselines across all poisoning rate settings, further solidifying its robustness and efficiency in backdoor defense.

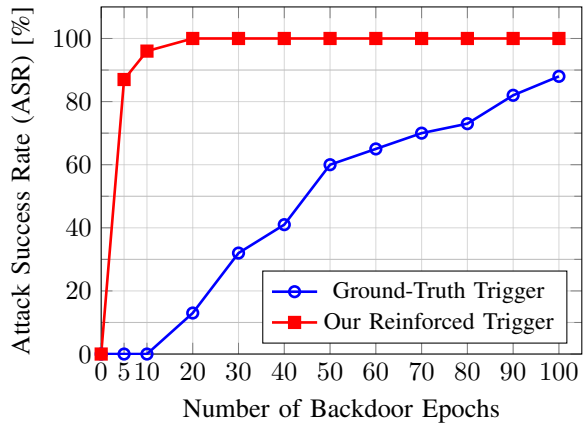
\begin{figure}[!t] 
\centering
\begin{tikzpicture}
\begin{axis}[
    width=0.9\columnwidth, 
    height=0.7\columnwidth, 
    xlabel={Number of Backdoor Epochs},
    ylabel={Attack Success Rate (ASR) [\%]},
    xmin=0, xmax=105,
    ymin=0, ymax=110,
    xtick={0, 5,10,20,30,40,50,60,70,80,90,100},
    ytick={0,20,40,60,80,100},
    legend pos=south east,
    grid=both,
    major grid style={opacity=0.6},
    minor tick num=1,
    legend style={font=\small}
]

\addplot[color=blue,mark=o,line width=1pt] coordinates {
    (0,0) (5, 0) (10, 0) (20, 13) (30, 32) (40, 41) (50, 60) (60, 65)
    (70, 70) (80, 73) (90, 82) (100, 88)
};
\addlegendentry{Ground-Truth Trigger}

\addplot[color=red,mark=square*,line width=1pt] coordinates {
    (0,0) (5, 87) (10, 96) (20, 100) (30, 100) (40, 100) (50, 100)
    (60, 100) (70, 100) (80, 100) (90, 100) (100, 100)
};
\addlegendentry{Our Reinforced Trigger}

\end{axis}
\end{tikzpicture}
\caption{The effect of backdoor amplification over different backdoor training epochs.}
\label{fig:backdoor_asr}
\end{figure}

\begin{figure}[h!]
	\centering
	\includegraphics[scale=0.16]{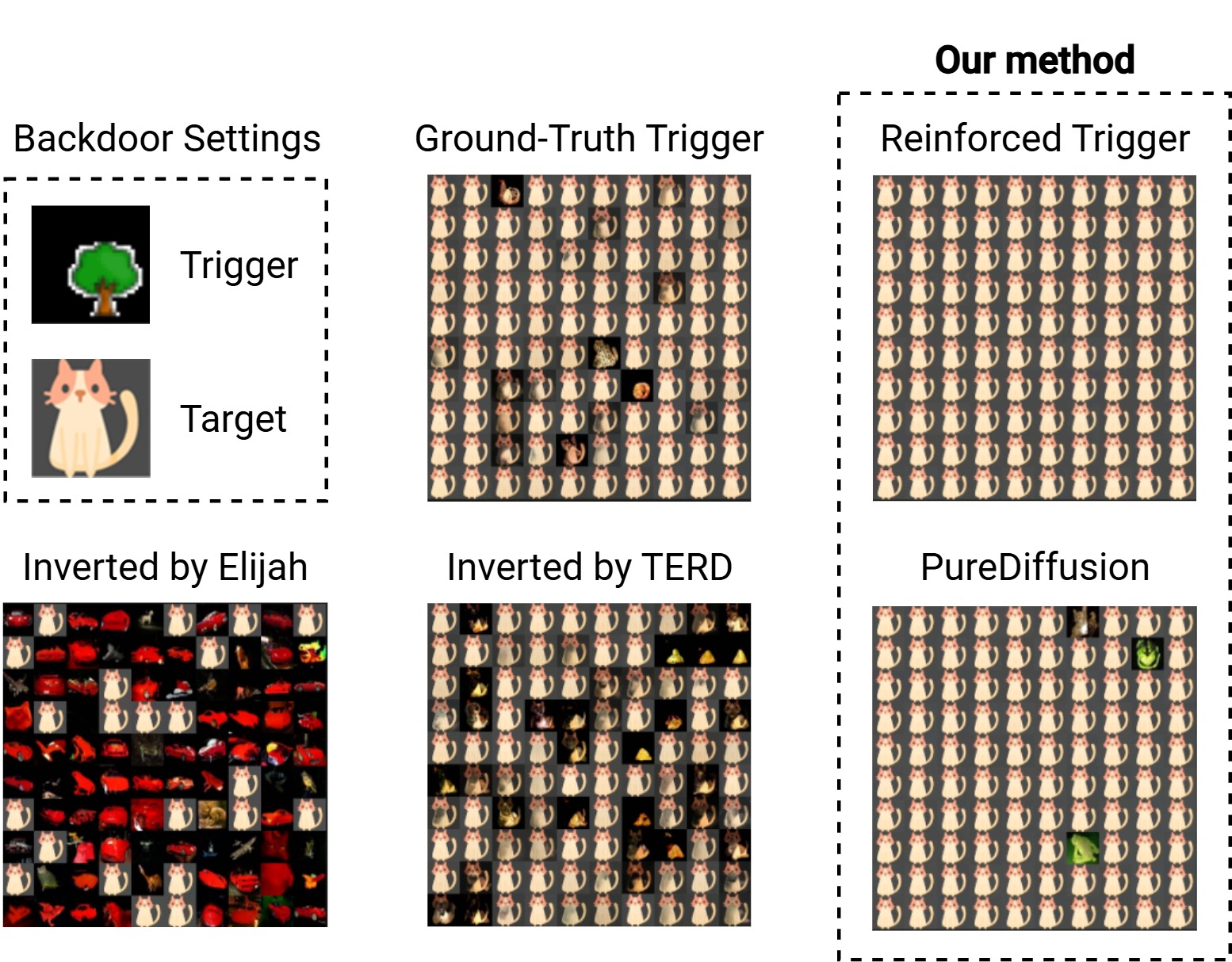}
	\caption{A visualization showing the efficiency of our method in both trigger inversion and trigger reinforcement for backdoor amplification.}
	\label{figure:trigger-ASR-visualization}
\end{figure}

\subsection{Performance of Backdoor Amplification}
As discussed earlier, after completing PureDiffusion's backdoor amplification, we obtain the reinforced trigger ${\boldsymbol{\delta}}^r$.
To evaluate the performance improvement of ${\boldsymbol{\delta}}^r$ over the original trigger ${\boldsymbol{\delta}}$, we conducted experiments on 11 diffusion models, each backdoored with a varying number of training epochs ranging from 5 to 100 epochs. The results, depicted in Fig.~\ref{fig:backdoor_asr}, highlight the significant advantage of our backdoor amplification technique. The reinforced trigger ${\boldsymbol{\delta}}^r$ achieves an ASR of 100\% after just 20 backdoor training epochs. In contrast, the original trigger ${\boldsymbol{\delta}}$ requires extended training, reaching only about 90\% ASR even after 100 epochs. Remarkably, our reinforced trigger achieves an equivalent ASR to that of the original trigger trained for 100 epochs, with the backdoor model trained for just 5 epochs. As optimizing the trigger is much faster than training backdoor models, utilizing PureDiffusion for backdoor amplification enables more efficient backdoor attacks, achieving almost 100\% ASR with over 20× faster training compared to traditional approaches.

Fig.~\ref{figure:trigger-ASR-visualization} presents a visual comparison of samples generated by the same backdoored diffusion model using various triggers. These include the ground-truth trigger, the triggers inverted by Elijah, TERD, and PureDiffusion, as well as the reinforced trigger from our framework. Each trigger was employed to generate 100 samples, and the results reveal the following: The triggers inverted by Elijah and TERD exhibit significantly lower ASR compared to the ground-truth trigger. In contrast, the trigger inverted by PureDiffusion achieves an ASR higher than the ground-truth trigger, demonstrating the superior inversion capability of our framework. Most notably, the reinforced trigger generated by PureDiffusion yields a 100\% backdoor target, highlighting its exceptional performance in amplifying backdoor attacks. This visualization emphasizes the efficacy of PureDiffusion in both trigger inversion and backdoor amplification, setting it apart from existing methods.

\begin{table}[]
\centering 
\caption{The impact of each loss function in trigger inversion.}
\label{table:ablation-trigger-inversion}
\begin{tabular}{@{}c|cc|c@{}}
\toprule
Performance Aspects & \multicolumn{2}{l|}{Backdoor Efficiency} & \multicolumn{1}{l}{Trigger Fidelity} \\ \midrule
Trigger Inversion Loss & SIM $\downarrow$ & ASR $\uparrow$ & L2D $\downarrow$ \\ \midrule
$L_{MDS}$ & \underline{0.00069} & \underline{92.75} & 29.82 \\
$L_{DC}$ & 0.00125 & 72.89 & \underline{19.33} \\
$L_{MDS}$ + $L_{DC}$ & \textbf{0.00062} & \textbf{93.12} & \textbf{18.26} \\ \bottomrule
\end{tabular}
\end{table}

\begin{table}[]
\centering
\caption{The impact of each backdoor detection strategy under specialized attack scenarios.}
\label{table:ablation-backdoor-detection}
\begin{tabular}{@{}c|ccc|ccc@{}}
\toprule
Attack Strategy & \multicolumn{3}{c|}{Multiple Targets} & \multicolumn{3}{c}{Invisible Trigger} \\ \midrule
Detection Strategy & ACC & TPR & TNR & \multicolumn{1}{l}{ACC} & \multicolumn{1}{l}{TPR} & \multicolumn{1}{l}{TNR} \\ \midrule
Generation-Based & 66.05 & 52.46 & 93.22 & \underline{97.08} & \underline{96.52} & \underline{98.20} \\
Trigger-Based & \underline{97.66} & \underline{97.11} & \underline{98.76} & 46.73 & 23.65 & 92.89 \\
Combined & \textbf{98.25} & \textbf{97.92} & \textbf{98.90} & \textbf{97.76} & \textbf{97.02} & \textbf{99.23} \\ \bottomrule
\end{tabular}
\end{table}

\subsection{Ablation Study}
We analyze the impact of each component in our framework on overall performance. First, we evaluate the role of the proposed loss functions, $L_{MDS}$ and $L_{DC}$, by removing each during trigger inversion. As shown in Table~\ref{table:ablation-trigger-inversion}, $L_{MDS}$ enhances backdoor strength, yielding high ASR and low SIM, ensuring consistent generation of backdoor targets. Meanwhile, $L_{DC}$ improves trigger fidelity, making the inverted trigger more similar to the original. Combining both losses achieves the best balance between fidelity and backdoor effectiveness.

Next, we examine the effectiveness of our two backdoor detection methods, generation-based and trigger-based, against advanced attack strategies. The first strategy introduces multiple backdoor targets (three in our experiments), while the second constrains trigger magnitude to resemble Gaussian noise. As shown in Table~\ref{table:ablation-backdoor-detection}, generation-based detection struggles with multi-target attacks, as the diversity of backdoor targets increases SIM scores beyond the detection threshold. Conversely, trigger-based detection is less effective against near-Gaussian triggers, as their distribution closely resembles benign noise. By combining both methods, our framework achieves near-perfect detection accuracy, even against these advanced attacks.

\begin{figure}[htbp]
\centering
\begin{tikzpicture}
    \begin{axis}[
        xlabel={Maximum number of denoising steps},
        ylabel={Time complexity (seconds)},
        ylabel style={blue}, 
        ymin=0, ymax=40, 
        axis y line*=left, 
        xtick={1, 10, 20, 30, 40, 50},
        xmin=-3, xmax=55,
        width=7.5cm, height=6cm,
        bar width=5pt, 
        ybar,
        legend style={at={(0.5, 1.2)}, anchor=north, legend columns=-1}
    ]
        \addplot[
            ybar,
            fill=blue!30,
            draw=blue,
            bar shift=0pt
        ] coordinates {
            (1, 1.7)
            (10, 7.97)
            (20, 14.95)
            (30, 23.25)
            (40, 34.46)
            (50, 37.02)
        };
    \end{axis}

    \begin{axis}[
        ylabel={Performance (ASR)},
        ylabel style={red}, 
        ymin=0, ymax=100, 
        axis y line*=right, 
        axis x line=none, 
        xtick={1, 10, 20, 30, 40, 50},
        xmin=-3, xmax=55,
        width=7.5cm, height=6cm,
        legend style={at={(0.5, 1.35)}, anchor=north, legend columns=-1}
    ]
        \addplot[
            red,
            mark=*,
            sharp plot
        ] coordinates {
            (1, 11.34)
            (10, 18.55)
            (20, 21.78)
            (30, 43.92)
            (40, 71.47)
            (50, 82.33)
        };
    \end{axis}
\end{tikzpicture}
\caption{Complexity and performance of trigger inversion when changing the maximum number of denoising timesteps used in computing gradient descent.}
\label{fig:complexity-analysis}
\end{figure}
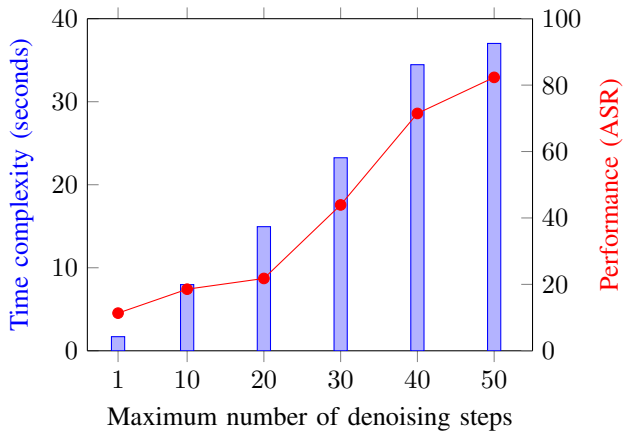

\subsection{Sensitivity and Complexity Analysis}
While previous sections highlight PureDiffusion’s strong performance across various scenarios, this section examines its complexity and sensitivity. As discussed earlier, we apply the losses $L_{MDS}$ and $L_{DC}$ only during the first $N$ denoising steps to balance defense robustness with computational efficiency. As the backward step $t$ for trigger inversion is sampled from 0 to $N$, the average number of steps to compute $\mathbf{x}_t$ is $N/2$.
We denote the numbers of iterations for $L_{MDS}$ and $L_{DC}$ are $n_{MDS}$ and $n_{DC}$, respectively. As a result, the computational complexity for trigger inversion is $O(\frac{N}{2}(n_{MDS}+n_{DC}))$.

We adjust $N$ to analyze its impact on the inverted trigger’s ASR and the time required for trigger inversion.
Fig.~\ref{fig:complexity-analysis} presents our results on an NVIDIA GeForce RTX 4090 GPU. As $N$ increases, ASR improves significantly, as the model gains more information about the trigger shift over multiple steps. However, this also extends the computational graph, increasing both runtime and resource consumption. With $N=50$, our trigger inversion algorithm takes approximately 40 seconds while achieving an ASR of over 80\%.

\section{Conclusion}\label{section:conclusion}
In this paper, we introduced PureDiffusion, a novel framework for countering backdoor attacks in diffusion models. By leveraging a two-stage trigger inversion method with two loss functions $L_{MDS}$ and $L_{DC}$, PureDiffusion effectively recovers backdoor triggers and achieves state-of-the-art performance in backdoor detection. Our experiments demonstrate that PureDiffusion significantly outperforms existing defense methods across various metrics, especially under challenging settings with difficult triggers and low backdoor poisoning rates. In addition to its defensive capabilities, PureDiffusion offers an unexpected yet impactful application: backdoor amplification. The proposed trigger reinforcement method optimizes triggers for adversarial use, enabling attackers to achieve near-perfect ASR with drastically reduced backdoor training time. By advancing the understanding of backdoor vulnerabilities and mitigation strategies, this work lays a foundation for future research in securing generative diffusion models.

\footnotesize{
\bibliographystyle{IEEEtran}
\bibliography{reference.bib}

\begin{thebibliography}{10}
\providecommand{\url}[1]{#1}
\csname url@samestyle\endcsname
\providecommand{\newblock}{\relax}
\providecommand{\bibinfo}[2]{#2}
\providecommand{\BIBentrySTDinterwordspacing}{\spaceskip=0pt\relax}
\providecommand{\BIBentryALTinterwordstretchfactor}{4}
\providecommand{\BIBentryALTinterwordspacing}{\spaceskip=\fontdimen2\font plus
\BIBentryALTinterwordstretchfactor\fontdimen3\font minus \fontdimen4\font\relax}
\providecommand{\BIBforeignlanguage}[2]{{%
\expandafter\ifx\csname l@#1\endcsname\relax
\typeout{** WARNING: IEEEtran.bst: No hyphenation pattern has been}%
\typeout{** loaded for the language `#1'. Using the pattern for}%
\typeout{** the default language instead.}%
\else
\language=\csname l@#1\endcsname
\fi
#2}}
\providecommand{\BIBdecl}{\relax}
\BIBdecl

\bibitem{ho2020denoising}
J.~Ho, A.~Jain, and P.~Abbeel, ``Denoising diffusion probabilistic models,'' in \emph{Proc. Int. Conf. Neural Inf. Process. Syst.}, vol.~33, 2020, pp. 6840--6851.

\bibitem{watson2021learning}
D.~Watson, W.~Chan, J.~Ho, and M.~Norouzi, ``Learning fast samplers for diffusion models by differentiating through sample quality,'' in \emph{Proc. Int. Conf. Learn. Representations}, 2021.

\bibitem{nichol2021glide}
A.~Nichol, P.~Dhariwal, A.~Ramesh, P.~Shyam, P.~Mishkin, B.~McGrew, I.~Sutskever, and M.~Chen, ``Glide: Towards photorealistic image generation and editing with text-guided diffusion models,'' in \emph{Proc. Int. Conf. Mach. Learn.}, 2021, pp. 16\,784--16\,804.

\bibitem{sinha2021d2c}
A.~Sinha, J.~Song, C.~Meng, and S.~Ermon, ``D2c: Diffusion-decoding models for few-shot conditional generation,'' in \emph{Proc. Int. Conf. Neural Inf. Process. Syst.}, vol.~34, 2021, pp. 12\,533--12\,548.

\bibitem{bao2022analytic}
F.~Bao, C.~Li, J.~Zhu, and B.~Zhang, ``Analytic-dpm: an analytic estimate of the optimal reverse variance in diffusion probabilistic models,'' in \emph{Proc. Int. Conf. Learn. Representations}, 2022.

\bibitem{austin2021structured}
J.~Austin, D.~D. Johnson, J.~Ho, D.~Tarlow, and R.~Van Den~Berg, ``Structured denoising diffusion models in discrete state-spaces,'' in \emph{Proc. Int. Conf. Neural Inf. Process. Syst.}, vol.~34, 2021, pp. 17\,981--17\,993.

\bibitem{hoogeboom2021argmax}
E.~Hoogeboom, D.~Nielsen, P.~Jaini, P.~Forr{\'e}, and M.~Welling, ``Argmax flows and multinomial diffusion: Learning categorical distributions,'' in \emph{Proc. Int. Conf. Neural Inf. Process. Syst.}, vol.~34, 2021, pp. 12\,454--12\,465.

\bibitem{li2022diffusion}
X.~Li, J.~Thickstun, I.~Gulrajani, P.~S. Liang, and T.~B. Hashimoto, ``Diffusion-lm improves controllable text generation,'' in \emph{Proc. Int. Conf. Neural Inf. Process. Syst.}, vol.~35, 2022, pp. 4328--4343.

\bibitem{savinov2021step}
N.~Savinov, J.~Chung, M.~Binkowski, E.~Elsen, and A.~v.~d. Oord, ``Step-unrolled denoising autoencoders for text generation,'' in \emph{Proc. Int. Conf. Learn. Representations}, 2021.

\bibitem{xu2023dream3d}
J.~Xu, X.~Wang, W.~Cheng, Y.-P. Cao, Y.~Shan, X.~Qie, and S.~Gao, ``Dream3d: Zero-shot text-to-3d synthesis using 3d shape prior and text-to-image diffusion models,'' in \emph{Proc. Int. Conf. Comput. Vis. Pattern Recognit.}, 2023, pp. 20\,908--20\,918.

\bibitem{truong2024text}
V.~T. Truong and L.~B. Le, ``Text-guided real-world-to-3d generative models with real-time rendering on mobile devices,'' in \emph{Proc. IEEE Wireless Commun. Netw. Conf.}\hskip 1em plus 0.5em minus 0.4em\relax IEEE, 2024, pp. 1--6.

\bibitem{poole2022dreamfusion}
B.~Poole, A.~Jain, J.~T. Barron, and B.~Mildenhall, ``Dreamfusion: Text-to-3d using 2d diffusion,'' \emph{arXiv:2209.14988}, 2022.

\bibitem{chen2020wavegrad}
N.~Chen, Y.~Zhang, H.~Zen, R.~J. Weiss, M.~Norouzi, and W.~Chan, ``Wavegrad: Estimating gradients for waveform generation,'' in \emph{Proc. Int. Conf. Learn. Representations}, 2020.

\bibitem{popov2021grad}
V.~Popov, I.~Vovk, V.~Gogoryan, T.~Sadekova, and M.~Kudinov, ``Grad-tts: A diffusion probabilistic model for text-to-speech,'' in \emph{Proc. Int. Conf. Mach. Learn.}, 2021, pp. 8599--8608.

\bibitem{tashiro2021csdi}
Y.~Tashiro, J.~Song, Y.~Song, and S.~Ermon, ``Csdi: Conditional score-based diffusion models for probabilistic time series imputation,'' in \emph{Proc. Int. Conf. Neural Inf. Process. Syst.}, vol.~34, 2021, pp. 24\,804--24\,816.

\bibitem{yan2021scoregrad}
T.~Yan, H.~Zhang, T.~Zhou, Y.~Zhan, and Y.~Xia, ``Scoregrad: Multivariate probabilistic time series forecasting with continuous energy-based generative models,'' \emph{arXiv:2106.10121}, 2021.

\bibitem{rasul2020multivariate}
K.~Rasul, A.-S. Sheikh, I.~Schuster, U.~Bergmann, and R.~Vollgraf, ``Multivariate probabilistic time series forecasting via conditioned normalizing flows,'' in \emph{Proc. Int. Conf. Learn. Representations}, 2020.

\bibitem{goodfellow2014generative}
I.~Goodfellow, J.~Pouget-Abadie, M.~Mirza, B.~Xu, D.~Warde-Farley, S.~Ozair, A.~Courville, and Y.~Bengio, ``Generative adversarial nets,'' in \emph{Proc. Int. Conf. Neural Inf. Process. Syst.}, vol.~27, 2014.

\bibitem{kingma2014vae}
D.~P. Kingma and M.~Welling, ``Auto-encoding variational bayes,'' in \emph{Proc. Int. Conf. Mach. Learn.}, 2014.

\bibitem{rezende2015variational}
D.~Rezende and S.~Mohamed, ``Variational inference with normalizing flows,'' in \emph{Proc. Int. Conf. Mach. Learn.}, 2015, pp. 1530--1538.

\bibitem{song2020denoising}
J.~Song, C.~Meng, and S.~Ermon, ``Denoising diffusion implicit models,'' in \emph{Proc. Int. Conf. Learn. Representations}, 2021.

\bibitem{song2019generative}
Y.~Song and S.~Ermon, ``Generative modeling by estimating gradients of the data distribution,'' in \emph{Proc. Int. Conf. Neural Inf. Process. Syst.}, vol.~32, 2019, pp. 11\,918--–11\,930.

\bibitem{song2020score}
Y.~Song, J.~Sohl-Dickstein, D.~P. Kingma, A.~Kumar, S.~Ermon, and B.~Poole, ``Score-based generative modeling through stochastic differential equations,'' in \emph{Proc. Int. Conf. Learn. Representations}, 2021.

\bibitem{truong2024attacks}
V.~T. Truong, L.~B. Dang, and L.~B. Le, ``Attacks and defenses for generative diffusion models: A comprehensive survey,'' \emph{arXiv:2408.03400}, 2024.

\bibitem{li2022backdoor}
Y.~Li, Y.~Jiang, Z.~Li, and S.-T. Xia, ``Backdoor learning: A survey,'' \emph{IEEE Trans. Neural Netw. Learn. Syst.}, vol.~35, no.~1, pp. 5--22, 2022.

\bibitem{wang2019neural}
B.~Wang, Y.~Yao, S.~Shan, H.~Li, B.~Viswanath, H.~Zheng, and B.~Y. Zhao, ``Neural cleanse: Identifying and mitigating backdoor attacks in neural networks,'' in \emph{Proc. IEEE Symp. Secur. Priv.}\hskip 1em plus 0.5em minus 0.4em\relax IEEE, 2019, pp. 707--723.

\bibitem{chen2018detecting}
B.~Chen, W.~Carvalho, N.~Baracaldo, H.~Ludwig, B.~Edwards, T.~Lee, I.~Molloy, and B.~Srivastava, ``Detecting backdoor attacks on deep neural networks by activation clustering,'' \emph{arXiv:1811.03728}, 2018.

\bibitem{an2024elijah}
S.~An, S.-Y. Chou, K.~Zhang, Q.~Xu, G.~Tao, G.~Shen, S.~Cheng, S.~Ma, P.-Y. Chen, T.-Y. Ho \emph{et~al.}, ``Elijah: Eliminating backdoors injected in diffusion models via distribution shift,'' in \emph{Proc. AAAI Conf. Artif. Intell.}, vol.~38, no.~10, 2024, pp. 10\,847--10\,855.

\bibitem{guan2024ufid}
Z.~Guan, M.~Hu, S.~Li, and A.~Vullikanti, ``Ufid: A unified framework for input-level backdoor detection on diffusion models,'' \emph{arXiv:2404.01101}, 2024.

\bibitem{sui2024disdet}
Y.~Sui, H.~Phan, J.~Xiao, T.~Zhang, Z.~Tang, C.~Shi, Y.~Wang, Y.~Chen, and B.~Yuan, ``Disdet: Exploring detectability of backdoor attack on diffusion models,'' \emph{arXiv:2402.02739}, 2024.

\bibitem{hao2024diff}
J.~Hao, X.~Jin, H.~Xiaoguang, and C.~Tianyou, ``Diff-cleanse: Identifying and mitigating backdoor attacks in diffusion models,'' \emph{arXiv preprint arXiv:2407.21316}, 2024.

\bibitem{mo2024terd}
Y.~Mo, H.~Huang, M.~Li, A.~Li, and Y.~Wang, ``Terd: A unified framework for safeguarding diffusion models against backdoors,'' \emph{arXiv:2409.05294}, 2024.

\bibitem{ronneberger2015u}
O.~Ronneberger, P.~Fischer, and T.~Brox, ``U-net: Convolutional networks for biomedical image segmentation,'' in \emph{Proc. Int. Conf. Med. Image Comput. Comput. Assist. Interv.}, 2015, pp. 234--241.

\bibitem{Chou2023CVPR}
S.-Y. Chou, P.-Y. Chen, and T.-Y. Ho, ``How to backdoor diffusion models?'' in \emph{Proc. Int. Conf. Comput. Vis. Pattern Recognit.}, June 2023, pp. 4015--4024.

\bibitem{chen2023trojdiff}
W.~Chen, D.~Song, and B.~Li, ``Trojdiff: Trojan attacks on diffusion models with diverse targets,'' in \emph{Proc. Int. Conf. Comput. Vis. Pattern Recognit.}, 2023, pp. 4035--4044.

\bibitem{chou2024villandiffusion}
S.-Y. Chou, P.-Y. Chen, and T.-Y. Ho, ``Villandiffusion: A unified backdoor attack framework for diffusion models,'' in \emph{Proc. Int. Conf. Neural Inf. Process. Syst.}, vol.~36, 2024.

\bibitem{struppek2023rickrolling}
L.~Struppek, D.~Hintersdorf, and K.~Kersting, ``Rickrolling the artist: Injecting backdoors into text encoders for text-to-image synthesis,'' in \emph{Proc. Int. Conf. Comput. Vis.}, 2023, pp. 4584--4596.

\bibitem{zhai2023text}
S.~Zhai, Y.~Dong, Q.~Shen, S.~Pu, Y.~Fang, and H.~Su, ``Text-to-image diffusion models can be easily backdoored through multimodal data poisoning,'' in \emph{Proc. ACM Int. Conf. Multimed.}, 2023, pp. 1577--1587.

\bibitem{pan2023trojan}
Z.~Pan, Y.~Yao, G.~Liu, B.~Shen, H.~V. Zhao, R.~R. Kompella, and S.~Liu, ``From trojan horses to castle walls: Unveiling bilateral backdoor effects in diffusion models,'' in \emph{Proc. Int. Conf. Neural Inf. Process. Syst.}, 2023.

\bibitem{wang2023stronger}
H.~Wang, Q.~Shen, Y.~Tong, Y.~Zhang, and K.~Kawaguchi, ``The stronger the diffusion model, the easier the backdoor: Data poisoning to induce copyright breaches without adjusting finetuning pipeline,'' in \emph{Proc. Int. Conf. Neural Inf. Process. Syst.}, 2023.

\bibitem{li2023learnable}
S.~Li, J.~Ma, and M.~Cheng, ``Learnable invisible backdoor for diffusion models,'' \emph{OpenReview}, 2023.

\bibitem{wang2025t2ishield}
Z.~Wang, J.~Zhang, S.~Shan, and X.~Chen, ``T2ishield: Defending against backdoors on text-to-image diffusion models,'' in \emph{ECCV}, 2025, pp. 107--124.

\bibitem{guo2024copyrightshield}
Z.~Guo, S.~Liang, A.~Liu, and D.~Tao, ``Copyrightshield: Spatial similarity guided backdoor defense against copyright infringement in diffusion models,'' \emph{arXiv:2412.01528}, 2024.

\bibitem{Vu2025ICC}
T.~V. Truong and L.~B. Le, ``Purediffusion: Using backdoor to counter backdoor in generative diffusion models,'' in \emph{Proc. Int. Conf. Commun.}, June 2025.

\bibitem{kullback1951information}
S.~Kullback and R.~A. Leibler, ``On information and sufficiency,'' \emph{Ann. Math. Stat.}, vol.~22, no.~1, pp. 79--86, 1951.

\bibitem{lilearnable}
S.~Li, J.~Ma, and M.~Cheng, ``Learnable invisible backdoor for diffusion models,'' \emph{Openreview}, 2024.

\bibitem{krizhevsky2009learning}
A.~Krizhevsky, G.~Hinton \emph{et~al.}, ``Learning multiple layers of features from tiny images,'' 2009.

\end{thebibliography}
}


 





\end{document}